\documentclass[a4paper,fleqn]{cas-dc} % Use the 'cas-sc' class for single-column

\usepackage[compress]{natbib}
\usepackage{times}
\usepackage{epsfig}
\usepackage{graphicx}
\usepackage{amsmath}
\usepackage{amssymb}
\usepackage{amsthm}
\usepackage{booktabs}
\usepackage{multirow}
\usepackage{pifont}
\usepackage{booktabs} % for professional tables
\usepackage{multirow}
\usepackage{svg}
\usepackage{xcolor} % For color
\usepackage{amsmath}
\usepackage{amssymb}
\usepackage{pifont}%
\usepackage{newunicodechar}
\newunicodechar{≥}{\ensuremath{\ge}}

\usepackage{graphicx}
\usepackage{float}
\usepackage{hyperref}
\usepackage{subcaption}
\usepackage{diagbox}

\usepackage{algorithm}
\usepackage{algpseudocode}
\usepackage{tabularx}

\def\tsc#1{\csdef{#1}{\textsc{\lowercase{#1}}\xspace}}
\tsc{WGM}
\tsc{QE}
\tsc{EP}
\tsc{PMS}
\tsc{BEC}
\tsc{DE}
\theoremstyle{definition}

  \newenvironment{itemize*}%  % Définition d'un nouvel environnement
    {\begin{itemize}%
      \setlength{\itemsep}{0pt}%
      \setlength{\parskip}{0pt}}%
    {\end{itemize}}

\begin{document}
\let\WriteBookmarks\relax
\def\floatpagepagefraction{1}
\def\textpagefraction{.001}

\newcommand{\comAM}[1]{{\color{red}#1}}

% Short title
\shorttitle{FedAgain: A Robust Federated Method for Kidney Stone Identification}

% Short author
\shortauthors{I. Reyes-Amezcua}

% Main title of the paper
\title [mode = title]{FedAgain: A Trust-Based and Robust Federated Learning Strategy for an Automated Kidney Stone Identification in Ureteroscopy}

\author[inst1]{Ivan Reyes-Amezcua}
\author[inst2,inst3]{Francisco Lopez-Tiro}
\author[inst3,inst4]{Clément Larose}
\author[inst3]{Christian Daul}
\author[inst1]{Andres Mendez-Vazquez}
\author[inst2]{Gilberto Ochoa-Ruiz}

\affiliation[inst1]{organization={Centro de Investigación y de Estudios Avanzados del IPN (CINVESTAV), Departamento de Ciencias Computacionales},
    city={Guadalajara}, postcode={45017},
    state={Jalisco}, country={Mexico}}

\affiliation[inst2]{organization={Escuela de Ingeniería y Ciencias, Tecnológico de Monterrey},
    city={Monterrey}, postcode={64849},
    state={N.L.}, country={Mexico}}

\affiliation[inst3]{organization={CRAN UMR 7039, Université de Lorraine and CNRS},
    %postcode={54516},
    city={Vand{\oe}uvre-lès-Nancy}, country={France}}

\affiliation[inst4]{organization={CHRU de Nancy-Brabois, Service d’urologie},
    city={Vandœuvre-lès-Nancy}, %postcode={54500}, state={Grand Est}, 
    country={France}}

\maketitle
\begin{abstract}
The reliability of artificial intelligence (AI) in medical imaging critically depends on its robustness to heterogeneous and corrupted images acquired with diverse devices across different hospitals which is highly challenging. Therefore, this paper introduces FedAgain, a trust-based Federated Learning (Federated Learning) strategy designed to enhance robustness and generalization for automated kidney stone identification from endoscopic images. FedAgain integrates a dual trust mechanism that combines benchmark reliability and model divergence to dynamically weight client contributions, mitigating the impact of noisy or adversarial updates during aggregation. The framework enables the training of collaborative models across multiple institutions while preserving data privacy and promoting stable convergence under real-world conditions. Extensive experiments across five datasets, including two canonical benchmarks (MNIST and CIFAR-10), two private multi-institutional kidney stone datasets, and one public dataset (MyStone), demonstrate that FedAgain consistently outperforms standard Federated Learning baselines under non-identically and independently distributed (non-IID) data and corrupted-client scenarios. By maintaining diagnostic accuracy and performance stability under varying conditions, FedAgain represents a practical advance toward reliable, privacy-preserving, and clinically deployable federated AI for medical imaging.

\end{abstract}

\section{Introduction}
\subsection{Medical Context}
Artificial Intelligence (AI) has made notable advances in various areas of medical imaging, supporting clinicians in disease detection, prognosis prediction, and treatment planning (\cite{li2024role}). For instance, AI systems have demonstrated strong capabilities in i) diagnosing pneumonia in pulmonary X-ray images and for tuberculosis screening (\cite{rajpurkar2017chexnet}), ii) detecting early signs of cardiovascular diseases from echocardiograms (\cite{ghorbani2020deeplearningechocardiogram}), iii) identifying malignant lesions in dermatology and histopathology (\cite{esteva2017dermatologist}) or iv) image modality transfer in endoscopy (\cite{Amaouche2025}). These examples illustrate how AI can handle ``visual'' tasks which are increasingly complex for human operators obser\-ving unimodal data, multimodal images or heterogeneous clinical data.

Beyond classification, classical applications of AI in medical imaging (see (\cite{litjens2017survey,liu2021advances,bian2025ai})) include organ or lesion segmentation, abnormality detection and localization, and tissue quantification (e.g., measuring tissue volumes, perfusion or vascular features).
However, the deployment of AI methods in the clinical practice is hampered by  inter-related challenges: \\
\noindent{- First,} the availability, diversity, and quality of annotated medical datasets remain a key bottleneck since many models are trained on curated research cohorts that are not representative of the full spectrum of clinical variability (device types, patient demographics, acquisition protocols, etc.). \\
\noindent{- Second,} deep-learning (DL) models often achieve excellent performance on internal test sets, but also present obstacles to trust, regulatory approval, and clinical acceptance (\cite{matus2020trustworthy}) due to their “black-box” nature which involves a lack of transparency on how predictions are made. Although Explainable AI (XAI) has emerged in response to this need (\cite{purwono2025xai}), such methods also face limitations in robustness, fidelity, and clinical usability (\cite{xu2024interpretable}). \\
\noindent{- Third}, AI models for medical image analysis are brittle towards (sometimes modest) perturbations (as noise, artifacts, device differences, patient motion) or domain shifts (due to new acquisition sites, populations, or acquisition devices) that can cause significant performance deterioration (\cite{hendrycks2019benchmarking}).

%These issues emphasize the urgent need for robust AI systems capable of maintaining performance under imperfect, corrupted, or unseen conditions.

% intro federated
In the field of endoscopy, training-data leakage and poor generalization remain common pitfalls (\cite{yoon2024domain}.) Many previous models in the literature have been trained and evaluated on single institutional datasets, which risks over-estimating performance due to a lack of diversity and potential data leakage (\cite{matta2024systematic}).  
For example, images from the same patient may appear in both training and test sets, leading models to overfit because of idiosyncrasies of imaging data from one single hospital (e.g., specific camera settings or patient demographics) and fail when deployed elsewhere. To mitigate these limitations, there is a growing push to leverage data from multiple hospitals, capturing the inherent heterogeneity of medical imaging. Although this strategy helps to reduce bias and improve generalization, it also introduces new challenges as inconsistency in data quality, heterogeneous label distributions, and even potentially corrupted or adversarial clients. Therefore, models must remain robust to a wider spectrum of distributional shifts, across different institutions (\cite{koccak2025bias}).

In the context of endoscopy, these challenges become even more acute. In addition to the variability introduced by the acquisition camera (resolution, contrast, light source color, etc.), endoscopic video-sequences of the gastric tract (gastroscopy (\cite{Zeteno2022Gastro}) and colonoscopy (\cite{Espinosa2024Colon})) and urinary  tract (cystoscopy (\cite{Ali2016Bladder}) and ureteroscopy (\cite{Villalvazo2023KidneyStones})) are encompassed by image sequences acquired in complex conditions, i.e., the movement of the endoscope is difficult to control due to deforming organs, leading to images affected by motion blur, specular reflections, occlusions, and variable illumination conditions depending on the viewpoint. AI models developed for endoscopic image analysis, such as automated lesion detection, segmentation of anatomical landmarks, navigation assistance, or kidney stone identification, must %therefore 
deal with strong domain shifts and the above mentioned perturbations (\cite{ali2022aiendoscopy}).

% Repetition : not necessary
%Recent advances in artificial intelligence for endoscopy aim to improve the diagnostic capability of physicians and procedural efficiency. However, these models are often brittle: small perturbations in the image (e.g., motion blur, specular highlights, sensor noise, or variable tissue appearance) can cause large errors in prediction \cite{ali2021endoscopicquality}.

In ureteroscopy, endoscopic image-based analysis plays a vital role in the identification and treatment of kidney stones. Ureteroscopy allows for the in-vivo visualization of kidney stones (i.e., visualization in the kidney calyces or ureters), and enables laser lithotripsy procedures (\cite{dretler1988laser}). AI-based approaches have been proposed for urinary stone detection, segmentation, 
%  MCA is not defined here : to early to speak about it. 
%\textcolor{blue}{and even Morpho-Constitutional Analysis (MCA)} 
and identification to determine the kidney stone composition from visual cues. However, the reliability of these algorithms fundamentally depends on their robustness to imaging perturbations, device heterogeneity, domain changes, and institutional variability (\cite{elbeze2022evaluation}). Thus, developing robust, explainable, and generalizable AI methods is essential to ensure the safe, trustworthy, and consistent performance of computer-assisted systems in endoscopic applications.

\subsection{Federated Learning in Healthcare}  
Federated Learning (Federated Learning) enables collaborative model training across multiple institutions without centralizing raw data, making it particularly well-suited to healthcare settings where privacy, institutional data fragmentation, and regulatory constraints (e.g., HIPAA\footnote{The Health Insurance Portability and Accountability Act  is an US law that limits the use of protected health information.} and GDPR\footnote{General Data Protection Regulation, UK and EU law to be considered when collecting personally identifiable information.}) are major concerns (\cite{rieke2020future, sheller2020federated}. Reviews of Federated Learning in medicine have shown that most works remain proof-of-concepts focused on open datasets and a limited number of institutions, while they only partially address real-world deployment challenges such as heterogeneity and robustness (\cite{xu2021federated}). For example, the systematic review in (\cite{antunes2022federated}) shows that only about 5.2\% of medical-Federated Learning studies involve multi-site real-life data collection. In the medical imaging domain, Federated Learning has been applied to tasks such as brain tumor segmentation, chest radiograph diagnosis, classification of the retinal fundus, and histopathology, demonstrating that performance comparable to that of central training can be achieved while preserving data locality (\cite{sheller2020federated, li2020multi, dou2021federated}). However, deployments in other fields (such as endoscopic kidney-stone imaging) remain rare. 

Despite these advances, a recent systematic survey notes that many strategies still leave long-term and client-specific problems largely unaddressed: Historical client behavior, persistent imaging biases, or repeated low-quality image uploads remain poorly handled (\cite{guan2024federated}). Autoencoder-based anomaly detectors help fill part of this gap by deploying a lightweight autoencoder at the edge, clients can minimize bandwidth and perform local outlier detection, flagging unreliable inputs or updates before uploading them (\cite{novoa2023fast, laridi2024enhanced}). However, most autoenco der work targets generic noise or blur in natural images, and rarely addresses the persistent domain-specific corruptions common in clinical imaging (for instance, endoscopic glare, specular highlights, and illumination drift), an issue repeatedly highlighted in medical-imaging Federated Learning reviews (\cite{kaissis2020secure, guan2024federated}.

The contribution in (\cite{reyes2025robust}) introduces the FedAgain algorithm, which integrates a dual-signal autoencoder mechanism into the federated loop to directly address such domain-specific anomalies. 
%
% \textcolor{blue}{poner imagen de trabajo previa, dodne se muestren la arquirecuta de los auto-encoders, tambien poner que es la novedad de este trabajo... NOT NECESSARY : TOO LONG WITH AN IMAGE}.
%
% 
Specifically, each client deploys a convolutional autoencoder to compute a reconstruction error \(r_k\) (anomaly score) in its local data and, after local update, the divergence \(d_k = \|w_{\rm server}^t - w_k^t\|_2\) of the client weights is calculated from the global model. The server then uses the combined signal \(\tau_k = \frac{1}{r_k \cdot d_k + \varepsilon}\) as a trust-weight to aggregate the update from the client. This turns anomaly detection into an active component of the federated training loop, rather than a passive filter.
Rather than using autoencoders only as a passive cleaning filter for classification and robust training tasks, it is essential that anomaly-detection signals be leveraged to improve the federated training itself. In other words, once a client or input is flagged as anomalous via high reconstruction error, this knowledge should inform the aggregation strategy, client weighting, or even the training schedule, thus turning the detection signal into an active component of the federated learning loop.

These considerations motivate the development of methods specifically designed for robust federated training in computer-aided diagnosis, which is the focus of the fol\-lo\-wing sections.

\subsection{Image Corruption in Federated Learning for Ureteroscopy}

In urology, kidney-stone imaging (ure\-te\-ros\-copy) involves several enduring challenges that directly affect the reliability of computer-assisted diagnosis and federated learning workflows. These challenges include specular reflections that are classical in endoscopy due to moisture-covered tissues, motion blur due to rapid camera or hollow organ movements, color drift across hospital imaging hardware and lighting setups, occlusions due to surgical instruments or biological debris, and variable illumination conditions or shadowing across the endoscopic video-sequences (\cite{lopez2021assessing, lopez2024vivo, Trinh2017IlluminationChanges}). In addition, in endoscopy over- or under-exposed images (\cite{GarciaVega2023ExposureCorrection}) frequently affect downstream tasks such as segmentation, detection, or feature extraction (\cite{reyes2024leveraging}) or hollow organ cartography, as in case of the bladder (\cite{Weibel2012BladderCartography}).

A corruption-aware Federated Learning system that integrates a pre-trained Convolutional Neural Network (CNN) along with two phases, learning parameter optimization, and federated robustness validation on corrupted data has started to tackle this domain-specific challenge (\cite{reyes2024leveraging}). The results reported by the latter emphasize that robust Federated Learning pipelines for medical imaging must not only account for statistical heterogeneity (non-IID client data, i.e., data distributions that are not independent and identically distributed across clients, leading to distributional shifts between local datasets) and systems heterogeneity, but also actively handle acquisition and hardware-specific impairments. These results lend strong support to the design of Federated Learning workflows that integrate both robust aggregation and client-side anomaly detection to achieve reliable, privacy-preserving medical diagnosis.

\subsection{Robustness to Image Corruptions}
%
% Next sentence is rather blabla and a repetion : but keep it if you want
%
Robustness in computer vision refers to the ability of a model to maintain a reliable performance even when confronted with image perturbations such as blur, noise, scene deformations, illumination changes or occlusions (\cite{hendrycks2019benchmarking, wang2023survey}). In medical imaging, robustness is not only desirable but essential, as degraded or Out-Of-Distributions (OOD) data can lead to diagnosis errors with potentially severe clinical consequences (\cite{koccak2025bias}). Endoscopic, radiologic, and histopathological images are inherently susceptible to these perturbations, which may arise from sensor limitations, motion during acquisition, or intra-operative environmental variability.

Recent research has begun to address robustness in distributed or federated applications. For example,{(\cite{fang2023robust}) proposed corruption-aware weighting mechanisms to mitigate the influence of corrupted local data on global model aggregation. However, such approaches have predominantly been validated on non medical image benchmarks (e.g., CIFAR-10-C, ImageNet-C) rather than on clinically relevant or intra-operative data. Consequently, the robustness of federated models in high-stakes medical applications, where data, inherently  diverse and noisy, remains sensitive to privacy and until now rather inadequately characterized.

Although both Federated Learning and robust learning have independently advanced, their convergence is still an emerging area of research. Existing robust-Federated Learning studies mainly focus on defending models against adversarial or malicious clients (\cite{pillutla2022robust}), whereas most medical-Federated Learning literature emphasizes privacy preservation, communication efficiency and data heterogeneity mitigation \cite{rieke2020future, sheller2020federated}. However, they often overlook robustness to naturally corrupted or degraded images. This disconnection is especially evident in urology, where endoscopic sequences frequently exhibit motion blur, specular, and occlusion by surgical tools or biological debris.

There is therefore a critical need for federated systems that jointly address three orthogonal, yet interdependent dimensions: i) patient privacy and regulatory compliance, ii) statistical and system heterogeneity across clinical sites, and iii) robustness to visual corruption and acquisition artifacts. 

% \textcolor{blue}{For me this text comes to early and is also probably repeated in section 2.} The proposed framework, FedAgain, integrates these aspects through a trust-weighted aggregation mechanism that leverages both benchmark reliability and divergence-based trust metrics to downweight unreliable clients. By doing so, FedAgain achieves resilient model convergence under challenging conditions, demonstrating improved stability and accuracy on both canonical benchmarks (MNIST, CIFAR-10) and multi-institutional ureteroscopic kidney-stone imaging datasets. This unified treatment of robustness, privacy, and heterogeneity offers a practical step toward safer and more trustworthy federated medical AI.

\section{Motivation and Contributions}

The primary motivation of this study is to advance collaborative learning across multiple hospitals while preserving privacy and ensuring that the resulting models are robust to real-world deployment conditions.
Federated Learning offers an attractive solution for multi-center training without centralizing sensitive patient data (\cite{mcmahan2017communication, kaissis2020secure}). Given that in a hospital setting, privacy regulations often prohibit the pooling of patient data, Federated Learning overcomes this by transmitting model updates instead of raw images (\cite{sheller2020federated}). Using Federated Learning, the proposed approach allows algorithms to learn from a larger and more diverse data set than any single institution could provide, thus improving generalization (\cite{ng2021federated}).  

However, training on distributed data alone does not guarantee robustness which is necessary to deal with corrupted data. 
%% Blabla and repetitions 
In real-world endoscopic procedures, conditions are often suboptimal, with image corruption being a common everyday scenario. 
Furthermore, common challenges include hardware limitations (e.g., older endoscope models), abrupt illumination changes, the presence of blood or stone dust, and limited computing resources on-site. Such factors can severely affect image quality and system reliability. These constraints motivate solutions that are lightweight, adaptive, and resilient. 

\noindent \textbf{Hypothesis.} This work posits that the intentional integration of dedicated robustness mechanisms, such as specialized model architectures and embedded client-side corruption awareness within the federated learning framework, is crucial for producing global models that are broadly applicable and resilient to a wide spectrum of image distortions (e.g., noise, blur, specularity, and color drift). The empirical results presented in the experimental section (Sec. \ref{Sec:Experimental}) serve to corroborate this hypothesis.

The contribution of this work lies on the introduction of a trust-based federated learning strategy that explicitly addresses the dual challenges of data privacy and robustness in medical imaging. By proposing FedAgain, a federated learning framework that dynamically weights client contributions based on benchmark reliability and model divergence, this work enhances robustness and generalization under heterogeneous, non-IID, and corrupted-client conditions. The proposed approach enables collaborative model training across multiple institutions while preserving data privacy and mitigating the impact of noisy or unreliable updates. To demonstrate the effectiveness and practical relevance of FedAgain, this paper focuses on the automated identification of kidney stone types from endoscopic ureteroscopy images, a clinically relevant task that exemplifies the challenges of real-world, safety-critical federated medical imaging applications.

%%% \subsection{Contributions}   REALLY JUST ONE SUBSECTON : 2.1 without 2.2
%% makes no sens
%This paper makes the following contributions:
The following contributions are defended in this work: 
\begin{itemize*}
    \item \textbf{Robust Federated Learning for Kidney Stone Imaging:} To the authors’ knowledge, this is the first federated learning approach for the in-vivo recognition of the kidney stone type that explicitly addresses the robustness of the model under image corruption and client unreliability. The proposed framework enables collaborative training to preserve privacy across multiple institutions using only client updates (no raw data shared).
    \item \textbf{Trust-Weighted Aggregation through FedAgain:} This work expands {\bf FedAgain}, a simple yet effective aggregation rule that computes a per-round client trust weight based on two signals: The benchmark error from a client on the incoming global model, and the divergence of its update from the global model.
This mechanism down-weights unreliable or corrupted clients without explicit outlier detection or attack-specific tuning, supporting resilience in non-IID and corrupted federated settings.
    \item \textbf{Wide-Scope Evaluation Framework:} This contribution provides a reproducible PyTorch codebase implementing standard federated algorithms (\textsc{FedAvg} (\cite{mcmahan2017communication} and \textsc{FedProx} (\cite{li2020federated})), robust aggregation baselines (coordinate-wise median (\cite{fang2022robust}), Bulyan (\cite{guerraoui2018hidden})), and FedAgain with support for non-IID Dirichlet splits. This work evaluates using a comprehensive suite of seven stress-tests that vary the corruption rates from 10\% up to 50\%, severity, and client counts, covering both public benchmarks (MNIST, CIFAR-10) and three clinical kidney-stone imaging datasets.

    \item \textbf{Empirical Validation of Robustness Gains:} Through experiments under realistic perturbations (noise, blur, contrast changes), it is shown that FedAgain outperforms both centralized and local baselines in terms of precision and robustness on kidney stone classification, and also maintains strong performance on standard image benchmarks. Ablation studies isolate the contributions of the trust-weighting scheme and federated training, demonstrating that both are essential for improved reliability in heterogeneous and corrupted-client federated scenarios.
\end{itemize*}

%%% This is already a conclusion : comes too early
Together, these contributions advance towards a clinically viable, privacy-compliant federated AI assistant for an in-vivo recognition of kidney stones that is robust to both data heterogeneity and image quality degradation.

%\section{Method <- \textcolor{blue}{Better title must be found}}
\section{Towards a new Trust-Based Federated Learning Method}
\subsection{Preliminaries on Federated Learning\label{sec:preliminaries}}  
\begin{figure*}[h]
    \centering 
    \includegraphics[width=0.7\textwidth]{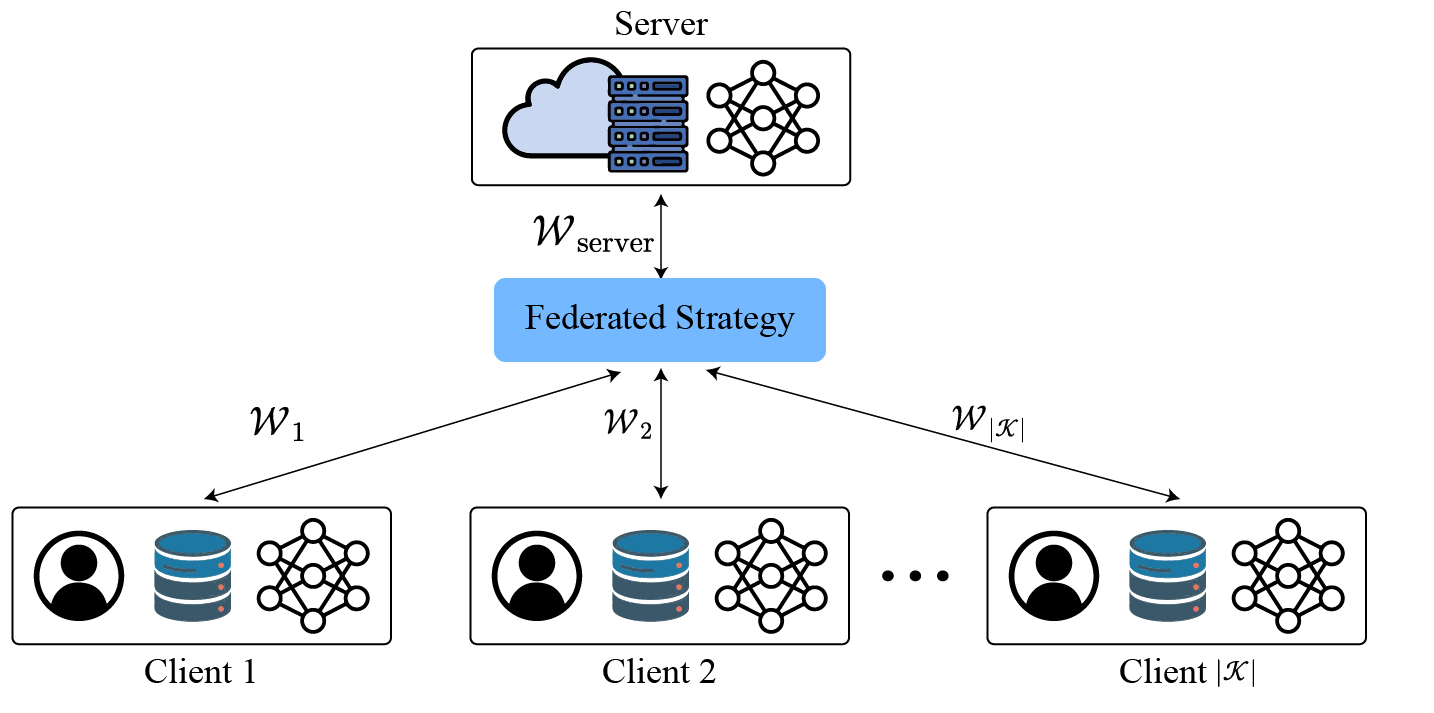} 
    \caption{Overview of the federated learning paradigm. At round $t$, each client $k \in [1, |\mathcal{K}|]$ (e.g., a public hospital or a private clinic) trains a local model on its private dataset $\mathcal{TD}_k$ and transmits only the resulting sets $\mathcal{W}_k^{(t)}$ of model parameters $w_{k, j}^{(t)}$ to the central server. The server applies a federated strategy, such as weighted averaging (see Eq. \eqref{eq:2}), to aggregate these updates and produce a new server model weight set $\mathcal{W}_{\textrm{server}}^{(t+1)}$. This updated model is then redistributed to all clients, enabling thus to exploit collaborative learning across heterogeneous and decentralized data sources without exposing private client data. 
    \label{fig:federated_overview} 
    }
\end{figure*}

Collaborative training without centralizing raw data has emerged as a key paradigm in privacy-sensitive domains.  As sketched in Fig. \ref{fig:federated_overview} and detailed in the next paragraphs, 
Federated Learning enables multiple clients to jointly train a global model by exchanging only model updates (for instance, weights or gradients), while keeping their local data private (\cite{kairouz2021advances,li2019survey}).  
The standard algorithm used in this area is {\scshape FedAvg}, which involves calculating a weighted average of the local model updates during each communication round (\cite{mcmahan2017communication}). 

\medskip  
\noindent\textbf{Mathematical formulation.}  
% \textcolor{blue}{TO BE CHECKED} \\
%
Consider a set \(\mathcal{K}\) of clients indexed by $k$ = 1, 2, $\ldots$, $|\mathcal{K}| - 1$, $|\mathcal{K}|$. Let \(\mathcal{TD}_k\) and \(\mathcal{VD}_k\) respectively denote the training and validation datasets located at client \(k\). The amount of samples in these local training and validation sets are \(|\mathcal{TD}_k|\) and \(|\mathcal{VD}_k|\), respectively, and the total number of training samples \(|\mathcal{TD}_{tot}|\)spread over all clients is given by:
\begin{equation} \label{eq:1}
|\mathcal{TD}_{tot}| =  \sum_{k=1}^{k = |\mathcal{K}|}|\mathcal{TD}_k| \;\, \textrm{and} \;\, \alpha_k = \frac{|\mathcal{TD}_k|}{|\mathcal{TD}_{tot}|} \in [0, 1] \; \forall \; k 
\end{equation}
The models on the central server (without raw data) and on all clients $k$ (each with its local data) have parameter sets $\mathcal{W}_{server}$ and $\mathcal{W}_{k}$  including all the same amount of weights $|\mathcal{W}_{server}|$ = $|\mathcal{W}_{k}| \, \forall \, k$.   
At round \(t\), the server holds a global model with a set of  weights \(\mathcal{W}_{\rm server}^{\,(t)}\).  
Each client $k$ receives this global model and performs local training on its set \(\mathcal{TD}_k\) (for example \(E\) epochs of stochastic gradient descent) to obtain updated local weights \(w_{k, j}^{\,(t)}\), with $j\in [1, |\mathcal{W}_{k}|]$.  
The server then aggregates the updates of all groups of corresponding weights $j$ of the clients to form the next global model (i.e., that of round $t+1$). This aggregation is done with Eq. \eqref{eq:2} which is based on the definition of $\alpha_{k}$ given in Eq. \eqref{eq:1}.
\begin{equation} \label{eq:2}
    w_{{\rm server},\,j}^{\,(t+1)}
= \sum_{k=1}^{k=|\mathcal{K}|} \alpha_k \, w_{k,\,j}^{\,(t)} \quad \textrm{with} \quad
    \sum_{k=1}^{k=|\mathcal{K}|} \alpha_k = 1
\end{equation}
%
\begin{comment}
 \textcolor{red}{Equation \eqref{eq:1}, which expresses the core idea of the FedAvg strategy, is more precisely written in  Eq. \eqref{eq:2}: }
\begin{equation} \label{eq:2}
w_{\rm server}^{\,(t+1)}
    = \sum_{k=1}^K \alpha_k \; w_k^{\,(t)}
    \,
    \textcolor{red}{\text{with}} \,
    \alpha_k \ge 0\ \textcolor{red}{\forall k} \, \textrm{and}
    \sum_{k=1}^K \alpha_k = 1.
\end{equation}
\end{comment}
%
Federated Learning enables to account for client-specific weighting schemes beyond simple sample‐size proportional weighting.

\medskip   
While {\scshape FedAvg} performs well under ideal IID data distributions, it is known to struggle when client data are heterogeneous (non-IID), client compute budgets differ (client computing heterogeneity), or when some clients may be corrupted or adversarial (\cite{li2019survey,mcmahan2017communication}).  
In response to these limitations, an increasing amount of research has sought to introduce more reliable aggregation and ensemble techniques. In this sense, more specialized strategies have been developed:
\begin{itemize}
  \item {\bf FedAvg (\cite{mcmahan2017communication}).} The canonical baseline strategy: Clients train locally on their private datasets $\mathcal{TD}_k$ and send their sets $\mathcal{W}_k^{(t)}$ of updated model parameters $w_{k,\,j}^{(t)}$ to the server in which the set $\mathcal{W}_\textrm{server}^{(t+1)}$ of parameters $w_{\textrm{server},j}^{\,(t+1)}$ is obtained by a weighted average of parameters \(w_k^{(t)}\), see Eq. \eqref{eq:2}. 
%
  %%     SAME FORMULA A SECOND TIME ? 
  %via a weighted average:
  %\[
  %  w_{\rm server}^{\,t+1}
  %  = \sum_{i=1}^K \frac{n_i}{N} \; w_i^t,
  %\]
  %where \(n_i\) is the number of samples at client \(i\) and \(N = \sum_i n_i\). 
  The strength of FedAvg lies in its simplicity and efficiency, but its weakness emerges in real-world federated scenarios with heterogeneity or adversaries.
\item {\bf FedProx (\cite{li2020federated})} is a heterogeneity-aware extension of FedAvg. At each communication round $t$, each client $k$ optimizes a proximal-augmented local objective $O_k^{(t)}$ as follows.
\begin{comment}
\[
 \;=\; F_k(w) \;+\; \frac{\mu}{2}\,\| w - w^{(t)} \|_2^2,
\]
\end{comment}
  %
  \begin{equation} \label{eq:3}
  % \textcolor{red}{O_k^{(t)} =} \min_{w\textcolor{red}{_k^{\,t}}} F_k(w\textcolor{red}{_k^{\,t}}) \;+\; \frac{\mu}{2}\,\|w\textcolor{red}{_k^{\,t}} - w_{\rm server}^{\,t}\|^2
  O_k^{(t)} = \min_{\mathcal{W}_k^{(t)}} F_k\left(\mathcal{W}_k^{(t)}\right) +\mu \hspace*{-1mm}\sum_{j=1}^{j=|\mathcal{J}|}\hspace*{-1.5mm}\left(w_{k,\, j}^{(t)} - w_{\textrm{server},\, j}^{(t)}\right)^2
  \end{equation}
In Eq. \eqref{eq:3}, the data-term $F_k\left(\mathcal{W}_k^{(t)}\right)$ represents the empirical risk  of client $k$ which is computed over its local dataset of weights. The proximal regularization term penalizes deviations of the local model weights $w_{k,\, j}^{(t)} \in \mathcal{W}_k^{(t)}$ from their corresponding global model weights $w_{\textrm{server},\, j}^{(t)} \in \mathcal{W}_\textrm{server}^{(t)}$ to mitigate client drifts induced by data heterogeneity. 
%The factor $\tfrac{1}{2}$ is included for notational and computational convenience, as it cancels out during gradient computation, while 
%The squared norm ensures a smooth and convex regularization.

The results reported in (\cite{li2020federated}) suggest that this approach improves convergence behavior under non-IID data and system heterogeneity, requiring fewer communication rounds to reach a fixed performance threshold and exhibiting lower variability across rounds.

%
  % \item {\bf FedMedian (\cite{yin2018byzantine}).} This approach is based on a robust aggregation rule that replaces the server’s weighted means $w_{{\rm server},j}^{\,(t+1)}$ with a coordinate-wise \textcolor{blue}{I do not now what is a coordinate wise median. Median of what ? Median smoothing of parameters $w_i^t$ ?} median (or similar robust statistic) of client updates. \textcolor{red}{With this method,} the influence of \textcolor{blue}{extreme <- ??? or malicious <- not OK simply inappropriate?} updates is reduced. However, this robustness comes with trade-offs, such as slower convergence and elevated variance in \textcolor{blue}{benign-only ?? What do you mean with benign?} settings.

    % corrected text:
  \item {\bf FedMedian \cite{yin2018byzantine}.} 
This method employs a robust aggregation rule in which the server replaces the standard weighted mean aggregation with a \emph{coordinate-wise median} computed over client model updates. 
Specifically, for each model parameter dimension $j$, the server aggregates $\{w_{i,j}^{(t)}\}_{i=1}^K$ by taking their median to obtain $w_{{\rm server},j}^{(t+1)}$. 
This operation reduces the influence of \emph{outlier updates}, which may arise from faulty, highly skewed, or adversarial clients. 
However, this robustness typically introduces trade-offs, including slower convergence and increased variance even in \emph{non-adversarial} (benign) training settings.

  \item {\bf Bulyan (\cite{guerraoui2018hidden})} is a two-phase Byzantine-robust aggregation scheme: First, a trusted subset of client updates is selected (e.g., via a Krum-style rule), then a trimmed-mean or median is applied over this subset. Bulyan offers more robust theoretical assurances in adversarial environments, although it comes with increased complexity and frequently operates under limiting assumptions.
\end{itemize}

This evolution, from naive averaging to heterogeneity-adapted optimization and robust statistics, reflects the maturation of Federated Learning research. Indeed, Federated Learning evolved from enabling privacy-preserving collaboration toward addressing heterogeneity, adversarial robustness, and generalization in real-world settings. %In this discussion ???, we <- not we proceed by 
This contribution focuses on robust federated training tailored for medical image analysis, a field where domain shifts and client heterogeneity are notably significant.
% \comAM{SOMETHING LIKE As we can see, the two previous problems in Federated Learning can be seen as a change in likelihood of distribution, thus the proposed FedAgain uses the information from such shifts to recalculate the likelihood at each stage of training to deal with the variances produced by changes in domain and clients.}
% \textcolor{blue}{aqui deberias seguir con un texto o sub-seccion de como tu propuesta construye sohre estos trabakjos porevios}. 

\subsection{Data Partitioning and Heterogeneity Simulation}
\label{subsec:data_partitioning}

A central challenge in Federated Learning is handling data heterogeneity across clients. In 
healthcare applications, each medical institution typically owns data generated under specific imaging devices, acquisition protocols, or patient demographics. This leads to variations in both data quantity and label distribution across sites, often resulting in \emph{non-identically distributed} (non-IID) client datasets, where the local data distributions differ across clients. Such non-IID data can significantly affect the efficiency of global aggregation, leading to slower convergence, client drift, and suboptimal generalization \cite{guan2024federated}.

 \emph{Dirichlet-based data partitioning} has become a standard approach in Federated Learning simulations to systematically study and quantify the effects of heterogeneity. Given a central dataset $\mathcal{DS}$
 %%%%%%%%%%%%%
 %%%  DS is different from divergence D in next section ! 
 %%%%%%%%%%%%%%%%
 (see the inputs of Algorithm \ref{Algo:1}), $|\mathcal{DS}|$ training samples are distributed among $|\mathcal{K}|$ clients according to class-specific Dirichlet proportions, controlled by concentration parameters $\beta_k$ 
 %%%   \beta_k is different from \alpha_k in Eq. 2
 with $k\in [1, \mathcal{K}]$ (\cite{hsu2019nonIID, yao2022lightfed}). This sample distribution is initialized in step 1 of Algorithm \ref{Algo:1}. A small $\beta_k$-value induces strong label imbalance (each client predominantly holds a few classes), whereas a large concentration $\beta_k$ approaches an IID-like distribution. This mechanism allows for the simulation of various levels of non-IIDness to evaluate the robustness of Federated Learning strategies under realistic clinical scenarios, where some hospitals may specialize in rare medical conditions or have limited patient diversity.
%%% I really do not understand this sentence : to be removed 
%In healthcare imaging, heterogeneity extends beyond label skew to encompass variations in modality, 
% Really I do not see what this examples illustrate : to be removed : heteroneity in dedine was alreay explained.
%e.g., CT (Computerized Tomography), MRI (Magnetic Resonance Imaging), endoscopy, scanner calibration, illumination, and disease prevalence (\cite{kaissis2020secure}). 
Realistic simulations of client data should incorporate both statistical non-IID distributions and domain-specific corruptions to mimic operational deployment.
 
Data heterogeneity can be ensured when the experimental setup exhibits a non-identically distributed (non-IID) label scenario, in which each participating institution is primarily specialized in a limited subset of diagnostic categories, while still retaining a small number of samples from other classes. Therefore, in this contribution, each simulated client was designed to focus on two predominant pathological groups, reflecting the specialization observed medical centers (for example, hospitals focusing on specific kidney-stone types or using different endoscopes). 

During a Dirichlet-based data partitioning, values of $\beta_k = 0.5$ are classically selected for all components of concentration vector $\theta_c$. This moderately low red {$\beta_k$-}value introduces controlled heterogeneity, generating uneven but not extreme class distributions across clients. Such a configuration offers a balanced compromise: prevent artificially uniform conditions that would underestimate domain-specific variability, while avoiding excessive label imbalance that could hinder convergence and generalization  (\cite{hsu2019nonIID, yao2022lightfed, guan2024federated}). Consequently, this setup enables a realistic assessment of  Federated Learning robustness under clinically plausible inter-institutional variability.

\begin{algorithm}[tb]
\caption{Dirichlet-based Non-IID Partitioning for Federated Learning.\label{Algo:1}}
\noindent{\bf Input:} central dataset $\mathcal{DS}$ with $|\mathcal{DS}|$ training samples, $|\mathcal{C}|$ classes $c\in\{1,\ldots,|\mathcal{C}|\}$, $|\mathcal{K}|$ clients $k\in\{1,\ldots,|\mathcal{K}|\}$, and class-specific Dirichlet concentration vectors $\{\boldsymbol{\beta}_c\}_{c=1}^{|\mathcal{C}|}$.\\
\noindent{\bf Output:} \(|\mathcal{K}|\) disjoint data-subsets \(\mathcal{DSS}^{(1)},\dots,\mathcal{DSS}^{(|\mathcal{K}|)}\).

\begin{enumerate}
\item {\bf Class--specific Dirichlet proportions:}\\
{\bf For} each class $c$ {\bf do:}
\begin{equation}\label{eq:theta}
  \boldsymbol{\theta}_c \sim \mathrm{Dirichlet}(\boldsymbol{\beta}_c)
\end{equation}

\item \textbf{Initialize empty client partitions:}
\begin{equation}
  \mathcal{DSS}^{(k)} \gets \varnothing \;\;\;\forall\,k\in\{1,\dots,|\mathcal{K}|\}
\end{equation}

\item \textbf{Assign samples to clients:}\\
{\bf For} each training sample of class $c$ {\bf do:}
\begin{itemize}
  \item Sample $k \sim \mathrm{Categorical}(\boldsymbol{\theta}_c)$
  \item $\mathcal{DSS}^{(k)} \gets \mathcal{DSS}^{(k)} \cup \{\text{sample}\}$
\end{itemize}
\end{enumerate}
\end{algorithm}

Algorithm \ref{Algo:1} produces controllable non-IID distributions that reflect realistic the diversity among clinical sites while maintaining statistical consistency for evaluation. When combined with corruption-aware validation (e.g., image artifacts or acquisition noise), these partitions enable robust benchmarking of federated aggregation algorithms under realistic multi-institutional healthcare conditions.

\begin{figure*}[t]
    \centering 
    \includegraphics[width=1\textwidth]{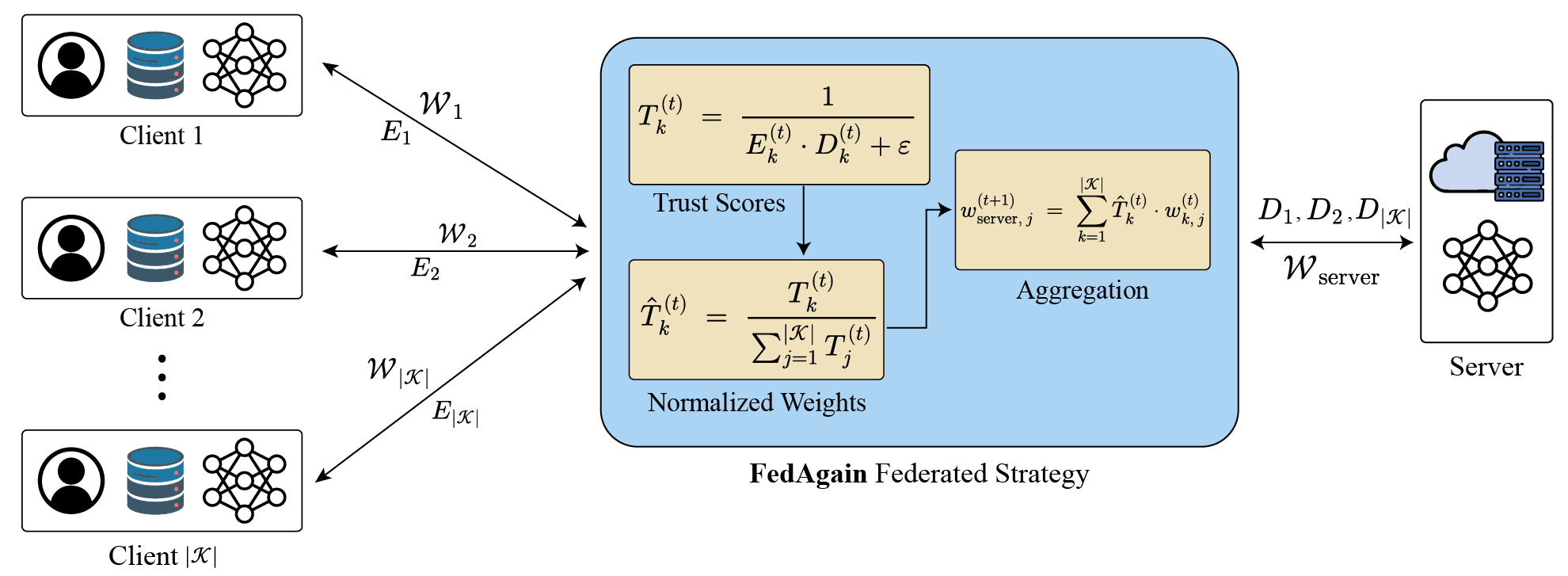} 
    \caption{Overview of the proposed \textbf{FedAgain} federated strategy. At round $t$, each client trains a local model $k\in [1, |\mathcal{K}|]$  on its private dataset $\mathcal{TD}_k$ and sends the resulting parameter set \( \mathcal{W}_k^{(t)} \), along with the benchmark error \( E_k^{(t)} \), 
    %and divergence \( D_k \)  <- Not true!
    to the server. The server computes a trust score \( T_k^{(t)}\) for each client, normalizes these trust scores to obtain weights \( \hat{T}_k^{(t)}\), and aggregates the local updated model weights $w^{(t)}_{k, \,j}$ to form the new global model $\mathcal{W}_\textrm{server}^{(t+1)}$. This trust-weighted aggregation scheme improves convergence robustness by down-weighting unreliable or corrupted clients. 
    % \newline \textcolor{blue}{Modification in this figure: 
    % \newline I) ON CLIENT SIDE: 1) Put three vertical dots ($\vdots$) between "Client 2" and "Client $k$", 2)  "Client $k$" must be  replaced by  "Client \textcolor{red}{$|\mathcal{K}|$}" 3) $W_k$ and $E_k$ and must be replaced by $\mathcal{W}_{|\mathcal{K}|}$ and $E_{|\mathcal{K}|}$ and 4) Divergences $D_1$,  $D_2$, and $D_{k}$ should be on server side. Remove them from client side.  
    % \newline II) IN THE "TRUST SCORE BOX" : $T_k$, $E_k$ and $D_k$ must be replaced by $T_k^{\textcolor{red}{(t)}}$, $E_k^{\textcolor{red}{(t)}}$ and $D_k^{\textcolor{red}{(t)}}$.
    % \newline III) IN THE "NORMALIZED WEIGHT BOX" : 1) $\hat{T}_k$, $T_k$ and $T_j$ must be replaced by $\hat{T}_k^{\textcolor{red}{(t)}}$ $T_k^{\textcolor{red}{(t)}}$ and $T_j^{\textcolor{red}{(t)}}$. 2) Upper sum limit $N$ must be replaced by \textcolor{red}{$|\mathcal{K}|$}. 
    % \newline IV) In the "AGGREGATION BOX": 1) sum over $k$ from 1 to \textcolor{red}{$|\mathcal{K}|$}  2) $w_{\textcolor{red}{\textrm{server}},\,\textcolor{red}{j}}^{(t+1)}$ instead of $w^{(t+1)}$, 3) $w_{k,\,\textcolor{red}{j}}^{\textcolor{red}{(t)}}$ instead of $w^{(t+1)}_k$ and 4) $\hat{T}_k$ must be replaced by $\hat{T}_k^{\textcolor{red}{(t)}}$
    % \newline V) ON SERVER SIDE: 1) $\textcolor{red}{\mathcal{W}}_{server}$ instead of  $w_{server}$, 2)  Divergences $D_1$,  $D_2$, and $D_{|\mathcal{K}|}$ should be on server side.}
    }
    \label{fig:fedagain_overview}
\end{figure*}

\subsection{Overview of the FedAgain Framework}
\label{sec:fedagain}

FedAgain is a robust Federated Learning strategy designed to improve reliability in decentralized training environments where data quality and client behavior may vary significantly. Its key innovation lies in combining two complementary dimensions of trust: (i) \emph{data quality}, assessed before local training, and (ii) \emph{model integrity}, assessed after local training (see Fig. \ref{fig:fedagain_overview}). These dual indicators are used to assign adaptive trust scores to clients during aggregation, mitigating the impact of unreliable or malicious participants.

\subsubsection{Federated Learning Workflow}
In a typical Federated Learning setting, a central server coordinates multiple multiple distributed clients, such as public hospitals, private clinics, or diagnostic devices, each with access to its own local dataset. The training process follows an iterative communication protocol:
\begin{enumerate}
    \item At round $t$, the server broadcasts the current set of weights \(\mathcal{W}^{(t)}_{\textrm{server}} \) of the global model to all clients. 
    \item At same round $t$, each client $k$ performs local training on its private dataset $\mathcal{TS}_k$ and produces local sets \(\mathcal{W}_k^{(t+1)}\) of updated weights.
    \item At the end of round $t$, the clients send their updated weights back to the server. The latter aggregates these weights into a new global model with weights \(\mathcal{W}^{(t+1)}_{\textrm{server}}\) that will be used at round $t+1$.
\end{enumerate}

Traditional methods such as FedAvg assume that all clients contribute equally to the aggregation of model updates, an assumption that is often violated in practical federated learning settings. Data quality across medical institutions can vary widely due to differences in imaging hardware, acquisition protocols, or annotation quality. Furthermore, certain clients may behave abnormally, either unintentionally (noisy data, limited convergence) or maliciously (poisoned updates). FedAgain explicitly addresses these issues by integrating a \emph{trust-based evaluation stage} into the standard Federated Learning pipeline.

% \textcolor{blue}{aqui seria bueno poner ya una figura concptual de fed again, con las posibles causas de inyeccion de ataques}. 

\subsubsection{Core Components of FedAgain}
FedAgain introduces two trust metrics computed during every training round $t$ (see Fig. \ref{fig:fedagain_overview}):

\begin{itemize}
    % \item \textbf{Benchmark Error \(E_k^{(t)}\)} is a measure of how well the current global model \(f(\cdot; \mathcal{W}^{(t)}_\textrm{server})\) performs on the local validation data \(\mathcal{VD}_k\) of client \(k\) before local training begins. This quantity reflects the representativeness and quality of the data distribution from each client and is defined as:
    % \begin{equation}
    %     E_k^{(t)} = \frac{1}{|\mathcal{VD}_k|} \sum_{i=1}^{i=|\mathcal{VD}_k|} \mathcal{L}\left(f(x_i; w^{(t)}_\textrm{server}),\, y_i\right)
    %     \label{eq:benchmark_error}
    % \end{equation}
    % where \(\mathcal{L}(\cdot)\) denotes the loss function (e.g., cross-entropy for classification), \(\mathcal{W}^{(t)_\textrm{central}}\) is the set of the current central model parameters, and \(|\mathcal{VD}_k|\) is the amount of samples in the client’s benchmark (validation) dataset. \textcolor{blue}{NOT SUFFICIENT: you must define $x_i$ and $y_i$, and explain in which situation the loss is (error) is minimized.
    % How is loss $\mathcal{L}(\cdot) = ?$ mathematically defined?}

    \item \textbf{Benchmark Error \(E_k^{(t)}\)} quantifies the performance of the current global model \(f(\cdot;\mathcal{W}^{(t)}_{\textrm{server}})\) on the local validation dataset \(\mathcal{VD}_k=\{(x_i,y_i)\}_{i=1}^{|\mathcal{VD}_k|}\) of client \(k\), prior to any local training at round \(t\). Here, $x_i$ denotes an input sample and $y_i$ its corresponding ground-truth label. This metric reflects the representativeness and quality of the client data distribution relative to the global model and is defined as
    \begin{equation}
        E_k^{(t)} = \frac{1}{|\mathcal{VD}_k|} \sum_{i=1}^{|\mathcal{VD}_k|} 
        \mathcal{L}\!\left(f(x_i;\mathcal{W}^{(t)}_{\textrm{server}}),\, y_i\right),
        \label{eq:benchmark_error}
    \end{equation}
    where $\mathcal{L}(\cdot)$ denotes a task-dependent loss function that measures the discrepancy between the model prediction and the ground-truth label. For classification tasks, $\mathcal{L}$ is typically instantiated as the cross-entropy loss, which is minimized when the predicted class probabilities match the true label distribution. Lower values of $E_k^{(t)}$ therefore indicate better alignment between the global model and the client’s validation data.

    \item \textbf{Parameter Divergence \(D_k^{(t)}\)} quantifies the deviation of the set of updated model parameters \(\mathcal{W}_k^{(t+1)}\) of client \(k\) from the set of central parameters \(\mathcal{W}^{(t)}_\textrm{server}\) after local training. This criterion assesses if the learning dynamics from a client align with the global optimization trajectory and is calculated as follows:
    \begin{equation}
        D_k^{(t)} = 
      %  \left\| w_k^{(t+1)} - w^{(t)}_{\textcolor{red}{server}} \right\|_2    <- THIS IS NOT OK FOR ME   
         \sqrt{ \sum_{j=1}^{j=|\mathcal{W}_{server}^{(t)}|} \left(w_{k,j}^{(t+1)} - w_{\textrm{server},j}^{(t)}\right)^2 }
        \label{eq:divergence}
    \end{equation}
    where \(|\mathcal{W}_{server}^{(t)}|\) is the total number of model parameters. \\
    \begin{comment}
    \textcolor{blue}{IMPORTANT FOR MATHEMATICAL NOTATION: I have a problem since the beginning of section 3: before this equation there was no difference between the model weight (or parameter) sets and the individual model parameter values. I suggest that we take following definitions: \\
    - The parameter/weight sets at round $t$ are called $\mathcal{W}_{server}^{(t)}$ for the server and $\mathcal{W}_{k}^{(t)}$ for client $k$. \\ 
    - The amount of model weights/parameters is $|\mathcal{W}_{server}^{(t)}|$= $|\mathcal{W}_{k}^{(t)}| \, \forall \, k$  \\
    - The $j^{th}$ model weight is $w_{server, j}^{(t)}$ and $w_{k, j}^{(t)}$ for the central server and client $k$, respectively.\\
    - In all figures, algorithms and everywhere in the text $W_k$ and $W_{server}$ make no sense (what is their meaning ?). For me, they must be replaced in all figures by  $\mathcal{W}_{server}^{(t)}$ and $\mathcal{W}_{k}^{(t)}$. \\
    IF YOU AGREE WITH THIS NOTATION, THESE DEFINITIONS SHOULD BE GIVEN AT BEGINNING OF SECTION 3 AND ALL THE TEXT, FIGURES AND ALGORITHMS MUST BE CORRECTED ACCORDINGLY.}  
    \end{comment}
\end{itemize}

% Both metrics are transmitted back to the server alongside the updated local model weights. The server then computes a composite \textbf{trust score} for each client:
% \begin{equation}
%     T_k^{(t)} = \frac{1}{E_k^{(t)} \cdot D_k^{(t)} + \varepsilon}
%     \label{eq:trust_score}
% \end{equation}
% where \(\varepsilon\) is a small real value to avoid null values at the denominator of Eq. \eqref{eq:trust_score}. 
% % MEANS NOTHING! : numerical stability. T
% High values of $T_k^{(t)}$ penalize clients exhibiting poor data quality (i.e., with high \(E_k^{(t)}\) values) or \textcolor{blue}{unstable <- means nothing! Be more precise} updates (i.e., with high values of  \(D_k^{(t)}\)), thereby reducing their influence during aggregation and enhancing overall robustness.

Both metrics are transmitted back to the server alongside the updated local model weights. The server then computes a composite \textbf{trust score} for each client:
\begin{equation}
    T_k^{(t)} = \frac{1}{E_k^{(t)} \cdot D_k^{(t)} + \varepsilon}
    \label{eq:trust_score}
\end{equation}
where \(\varepsilon > 0\) is a small constant introduced to prevent division by zero in Eq. \eqref{eq:trust_score}. High values of $T_k^{(t)}$ downweight clients whose updates are associated with high benchmark error \(E_k^{(t)}\) or exhibit large parameter deviations from the global model, as quantified by \(D_k^{(t)}\), thus reducing their influence during aggregation and improving robustness to noisy or corrupted client updates.

\subsubsection{System-Level Representation}
Figure~\ref{fig:fedagain_overview} outlines the complete \textbf{FedAgain} workflow. Each client \(k \in [1, \dots, |\mathcal{K}|]\) performs local training and transmits both its updated model parameter sets \(\mathcal{W}_k^{(t+1)}\) and performance criterion pair $\big(E_k^{(t)}, D_k^{(t)}\big)$ to the central server.

The server first computes a \emph{raw trust score} for each client according to Eq. \eqref{eq:trust_score}. 
%To ensure that the contributions of all clients sum to one, 
The server then normalizes these raw trust scores to obtain  final trust weights ranging from [0, 1]:
\begin{equation}
    \hat{T}_k^{(t)} = \frac{T_k^{(t)}}{\displaystyle\sum_{j=1}^{j = |\mathcal{K}|} T_j^{(t)}} \quad \textrm{with} \quad \sum_{k=1}^{k = |\mathcal{K}|} \hat{T}_k^{(t)} = 1.
    \label{eq:trust_normalized}
\end{equation}
The global aggregation is then performed for each central model weight $w^{(t+1)}_{\textrm{server}, j}$ as a trust-weighted combination of the corresponding ($j^{th}$) weights of all client $k$ updates:
\begin{equation}
    w^{(t+1)}_{\textrm{server},\,j} = \sum_{k=1}^{k = |\mathcal{K}|} \hat{T}_k \, w_{k,\,j}^{(t+1)}.
    \label{eq:trust_aggregation}
\end{equation}

Through this mechanism, FedAgain continuously adapts to the reliability of each participant by assigning higher weights to clients that exhibit both low benchmark error and low parameter divergence. As a result, the aggregation becomes resilient to natural data corruptions (e.g., low-quality or noisy data) and Byzantine manipulations (e.g., poisoned updates or adversarial parameter shifts), favoring 
%\textcolor{blue}{(forget stable! means nothing)} 
convergence in heterogeneous federated environments.

\subsection{Extending FedAgain Toward Corruption-Aware Robustness}
FedAgain can be interpreted as a dynamic reputation system embedded within the federated optimization loop, where the Federated Learning client reliability is continuously assessed based on dual trust dimensions, namely data quality and model consistency. Each client earns higher trust when:
\begin{itemize}
    \item Its data distribution is representative of the global population, resulting in a low benchmark error \(E_k^{(t)}\).
    \item Its local model updates remain consistent with the global optimization trajectory, leading to a low parameter divergence \(D_k^{(t)}\).
\end{itemize}
%
% Conversely, clients exhibiting high errors or divergences exert minimal influence on the aggregated model, effectively limiting the propagation of corrupted or unreliable updates. This adaptive weighting mechanism enables FedAgain to maintain \textcolor{red}{robust? not stable} convergence in complex federated environments characterized by non-IID data, limited supervision, and potential adversarial interference.

In contrast, clients with high errors or divergences have minimal influence on the aggregated model, effectively limiting the propagation of corrupted or unreliable updates. This adaptive weighting mechanism enables FedAgain to achieve \textbf{robust convergence} in complex federated environments characterized by non-IID data, limited supervision, and potential adversarial interference.

\paragraph{\bf Extension to Corruption Detection.} 
While our original formulation of FedAgain (}\cite{reyes2025robust}) focused primarily on trust-based aggregation, this work extends the framework toward \emph{corruption-aware robustness}. Specifically, client-side loss functions serve not only as optimization objectives, but also as diagnostic indicators of corruption or instability. Earlier formulations of FedAgain used accuracy as a proxy for reliability. However, accuracy provides only a coarse, discrete measure of correctness. In contrast, the loss-based metric such as cross-entropy captures the \emph{confidence} and \emph{calibration} of the prediction of a model, offering a continuous and differentiable reliability signal. 

%Mathematically, 
\paragraph{Example.}  
\label{example:FedAgain}
Suppose that the local model of client \(k\)  achieves a cross-entropy loss \(E_k^{(t)}\) of 0.25 and a divergence \(D_k^{(t)}\) of 0.8 and that for another client \(k'\) one have $E_{k'}^{(t)}$ = 0.5 and \(D_{k'}^{(t)} = 0.6\). Then their trust weights are:
\[
T_k^{(t)} = \frac{1}{0.25 \times 0.8 + \varepsilon} 
\quad \textrm{and} \quad
T_{k'}^{(t)} = \frac{1}{0.5 \times 0.6 + \varepsilon}.
\]
For \(\varepsilon {=} 10^{-3}\) the ratio of these two trust scores is given by: 
\[
\frac{T_k}{T_{k'}} \approx \frac{0.3}{0.2} = 1.5.
\]
In comparison to client $k'$, client \(k\) contributes \(50\%\) more strongly to the next global model update due to both the lower cross entropy loss and the moderate divergence.

The overall procedure of the proposed \textbf{FedAgain} framework is summarized in Algorithm \ref{Algo:2}. 
At the beginning of each communication round \(t\), the central server broadcasts the current global model parameter set \(\mathcal{W}^{(t)}_\textrm{server}\) to all participating clients. 
Each client \(k\in[1, |\mathcal{K}|]\) first evaluates the performance of the received model on its local validation data \(\mathcal{VD}_k\) by computing the benchmark error \(E_k^{(t)}\) defined in Eq. \eqref{eq:benchmark_error}. 
Subsequently, each client performs local training on its private dataset \(\mathcal{TD}_k\) to obtain an updated model parameter set \(\mathcal{W}_k^{(t)}\), which is, together with \(E_k^{(t)}\), sent back to the server. 

Upon receiving the local updates, the server computes parameter divergence \(D_k^{(t)}\) (defined in Eq. \eqref{eq:divergence}) for each client and derives  trust scores \(T_k^{(t)}\) (see Eq. \ref{eq:trust_score}) that jointly account for data quality and update consistency. 
These scores are normalized to form aggregation weights (see Eq. \eqref{eq:trust_normalized}). 
Finally, the global model is updated through a trust-weighted aggregation step (see Eq. \eqref{eq:trust_aggregation}) and the new parameter set \(\mathcal{W}^{(t+1)}_\textrm{server}\) of central server model is distributed to all clients for the next round $t+1$. 

This iterative process allows FedAgain to dynamically down-weight unreliable or corrupted clients to penalize their impact, while reinforcing contributions from trustworthy ones, thereby achieving robust convergence under heterogeneous and potentially adversarial federated learning conditions.

\begin{algorithm}[tb] 
\caption{FedAgain: server and client operations at round \(t\).%\textcolor{blue}{Question 1: exponent (t) for client set?} Is the client number not constant along $t$? $\mathcal{K}^{\textcolor{blue}{(t)?}}$ \\
%\textcolor{blue}{Question 2:} we probably need initial values for the divergences $D_k^{\textcolor{blue}{(t=1)}}$ at the first round (t=1)? \\
%\textcolor{blue}{Question 3: difference between steps 1 and 4?} 
\label{Algo:2}} 
\noindent{\bf Input:} Global parameter set $\mathcal{W}^{(t)}_\textrm{server}$, set $\mathcal{K}$ of $|\mathcal{K}|$ clients $k$, learning rate $\eta$ = 0.001 and constant $\varepsilon = 10^{-3}$ for Eq. \eqref{eq:trust_aggregation}. \\
\noindent{\bf Output:} Updated global model parameter set $\mathcal{W}^{(t+1)}_\textrm{server}$.
%\medskip
\begin{enumerate}
\item \textbf{Server → Clients:} \\ Distribute the current global model $w^{(t)}_{server}$ 
%\textcolor{blue}{and divergences $D_k^{\textcolor{blue}{(t-1) \, \textrm{OR} \, (t)?}}$}  <- NO this WAS AN ERROR FROM ME There was a confusion between local dataset D and divergence D! 
to all clients $k$ of set $\mathcal{K}$. 
\item \textbf{Client-side:} \\
{\bf For} all clients $k\in \mathcal{K}$ {\bf do}: 
\begin{itemize}
    \item Using Eq. \eqref{eq:benchmark_error}, compute benchmark error $E_k^{(t)}$ with the local validation dataset \(\mathcal{VD}_k\).
    \item Using the local dataset \(\mathcal{TD}_k\), perform training to update weight set $\mathcal{W}^{(t)}_{k}$: \\
    $\mathcal{W}_k^{(t)} = \textsc{LocalTrain}(\mathcal{W}^{(t)}_\textrm{server}, {\mathcal{TD}_k}, \eta)$.
    \item Send set $\mathcal{W}^{(t)}_{k}$ and benchmark error $E_k^{(t)}$ back to the server.
\end{itemize}
\item \textbf{Server-side (aggregation of global model weights):}
%\begin{itemize}
    {\bf For} all clients $k\in \mathcal{K}$ {\bf do}: 
    \begin{itemize}
      \item Using Eq. \eqref{eq:divergence}, compute divergence $D_k^{(t)}$ 
        \item  Using Eq. \eqref{eq:trust_score} and $E_k^{(t)}$ send by client $k$, compute trust score $T_k^{(t)}$ 
        \item Using Eq. \eqref{eq:trust_normalized}, compute normalized $\hat{T}_k^{(t)}$ 
        \item Aggregation: using Eq. \eqref{eq:trust_aggregation}, update all weights $w^{(t+1)}_{\textrm{server},\,j}$ of central model para\-meter set $\mathcal{W}^{(t+1)}_\textrm{server}$.   
%    \end{itemize}
\end{itemize}
\item \textbf{Go to step 1} ($t+1$ becomes $t$) 
%\textbf{Server → Clients:} \\
  %Broadcast $w^{(t+1)}_{server}$ and their \textcolor{red}{divergence $D_k^{\textcolor{blue}{(t)?}}$ ???} for the next $(t+1)$ round.
%
\end{enumerate}
\end{algorithm}

%\section{Datasets and Case Study}
\section{Federated Learning in Urology
\label{sec:datasets}}

\begin{figure*}[h]
    \centering 
    \includegraphics[width=1\textwidth]{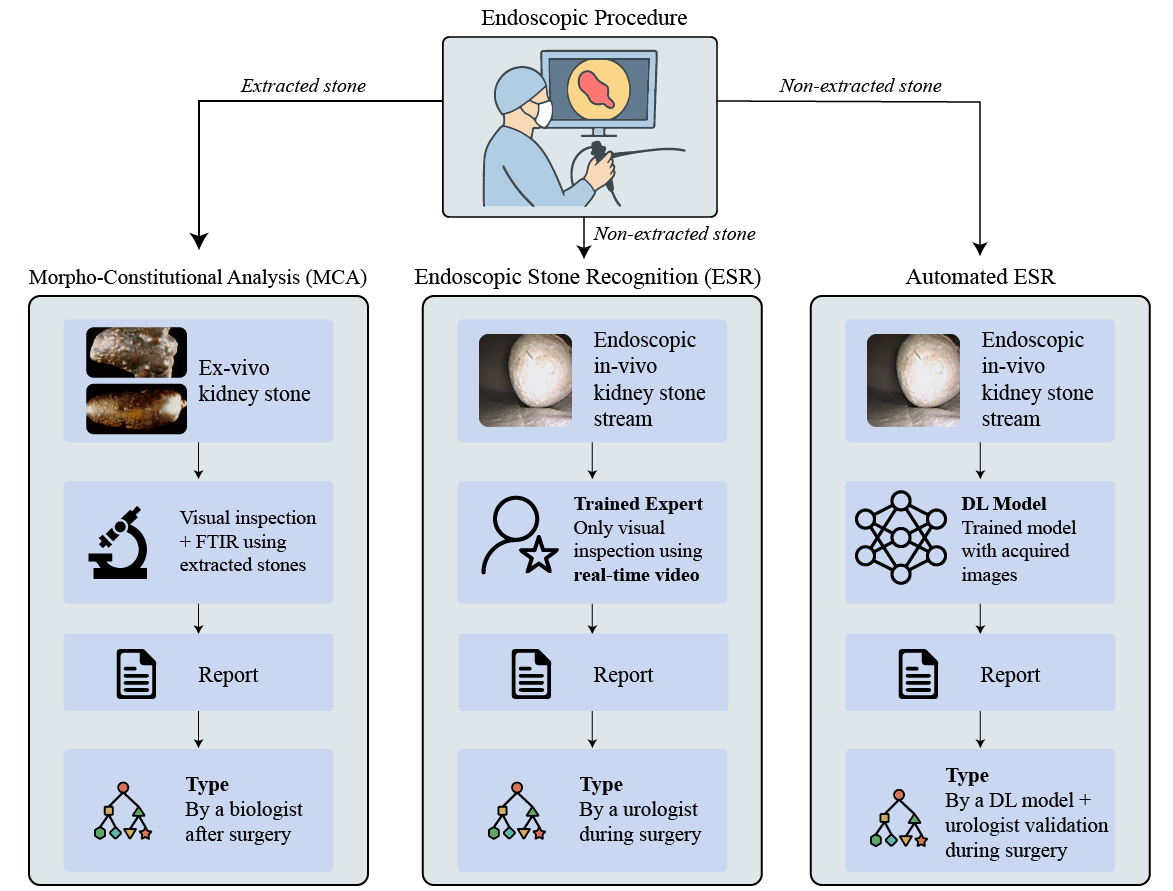} 
    \caption{Comparison of kidney stone identification methods. 
    Traditional MCA (left) relies on ex-vivo  visual inspection after extraction combined with FTIR, while endoscopic stone recognition (ESR, center) depends on the expert's visual assessment during the ureteroscopy. The proposed automated ESR approach (Title in the right colomn) integrates deep learning models for in-vivo stone type prediction.
    }
    \label{fig:mca_overview}
\end{figure*}

\subsection{Medical context: Kidney Stone Identification}

Kidney stone formation is a major public health issue, especially in developed countries, where up to \(10\%\) of the population can be affected by urolithiasis and in the United States the recurrence rate within five years can reach  \(40\%\) (\cite{kasidas2004renal}). 

Kidney stones (or renal calculi) are solid aggregations of crystalline material that form within the urinary tract as a result of mineral precipitation in urine. When their size increases beyond a few millimeters in diameter, the renal calculi can be blocked in the calyces of the kidneys or in the ureters, causing intense pain. Kidney stones are categorized into seven main types and 23 subtypes, each subtype being defined by a specific crystalline structure and biochemical composition. Their formation is influenced by a combination of intrinsic and extrinsic risk factors in the patient, including, on the one hand, genetic predisposition, age, body weight, or sex, and, on the other hand, environmental and physiological factors such as climate, lifestyle, metabolic comorbidities, or iatrogenic infections. The authors in  (\cite{cloutier2015kidney}) provide a comprehensive overview of kidney stone classifications, subtypes, and the etiological mechanisms underlying urolithiasis.

Identifying the subtype of kidney stones is essential for tailoring anti-relapse treatments and guiding prevention strategies. The Morpho-Constitutional Analysis (MCA) is a widely recognized renal calculi identification procedure (\cite{daudon2016comprehensive, corrales2021classification}). The MCA combines a visual inspection of the kidney stone fragments (extracted during an ureteroscopy) with a biochemical evaluation based on Fourier transform infrared spectroscopy (FTIR) analysis. Although MCA remains a gold standard, it is typically performed in biological laboratories of hospitals (ex-vivo analysis) which are usually able to provide the MCA-results only some weeks after the ureteroscopy, while for some stone types an immediate treatment is required or highly recommended.    
    
Endoscopic Stone Recognition (ESR), visually performed by a urologist during ureteroscopy itself (see Fig. \ref{fig:mca_overview}), could represent a medically interesting alternative. In-vivo stone type recognition would allow an immediate and appropriate treatment choice; for instance, a dusting procedure could be employed to destroy the urinary calculi instead of the classical fragmentation and extraction method, which typically lasts at least half an hour and involves infection risks. However, visual identification through ESR is highly operator-dependent, and only a limited number of experts (\cite{estrade2021ESR}) are able to reliably recognize kidney stone subtypes from endoscopic images. Most urologists do not have sufficient time in clinical practice to acquire this level of expertise.
 
% but it remains largely subjective and is complicated by challenging acquisition conditions such as changing illumination, motion blur, and noisy imagery (see Fig. \ref{fig:mca_overview}).

Recent research has increasingly focused on leveraging deep learning for the automatic image-based identification of kidney stones (\cite{Gonzalez2024MetricLearning, Turcotte2025comprehensiveAnalysis, Florez2025ProtoParts}).
%
%(\cite{corrales2021classification}). <- not appropriate here 
%
These methods have demonstrated promising performance in identifying kidney stone subtypes. However, most have been developed and validated on monocentric datasets. Such datasets often suffer from limited sample diversity, %\textcolor{blue}{inconsistent <- ???} image quality, and varying
mainly due to too constant acquisition protocols that restrict the assessment of generalizability of the models. Consequently, while the models can achieve high accuracy 
%\textcolor{blues}{under controlled laboratory conditions, <- makes no sens!}
their reliability and robustness tend to degrade when deployed in  clinical environments characterized by varying acquisition protocols and patient populations .

A Federated Learning framework presents a compelling alternative to overcome these limitations. 
% REPETITION AND BLABLA
%Federated Learning enables multiple institutions to collaboratively train a shared model without the need to exchange sensitive patient data, thereby preserving privacy and complying with healthcare regulatory requirements such as HIPAA and GDPR. More importantly
Federated Learning provides a natural setting for investigating and improving the robustness of a model across diverse and distributed data sources. By exposing the global model to the inherent variability of each institution’s imaging pipeline,  Federated Learning can foster the development of models that better generalize to unseen clinical conditions and resist common image corruptions or domain shifts.

\subsection{Kidney Stone Datasets}
\label{sec:datasets2}

\begin{figure*}[h]
    \centering
    \subfloat[Dataset A -- Handheld CCD camera]{
    \label{fig:dataseta}\includegraphics[width=0.485\linewidth]{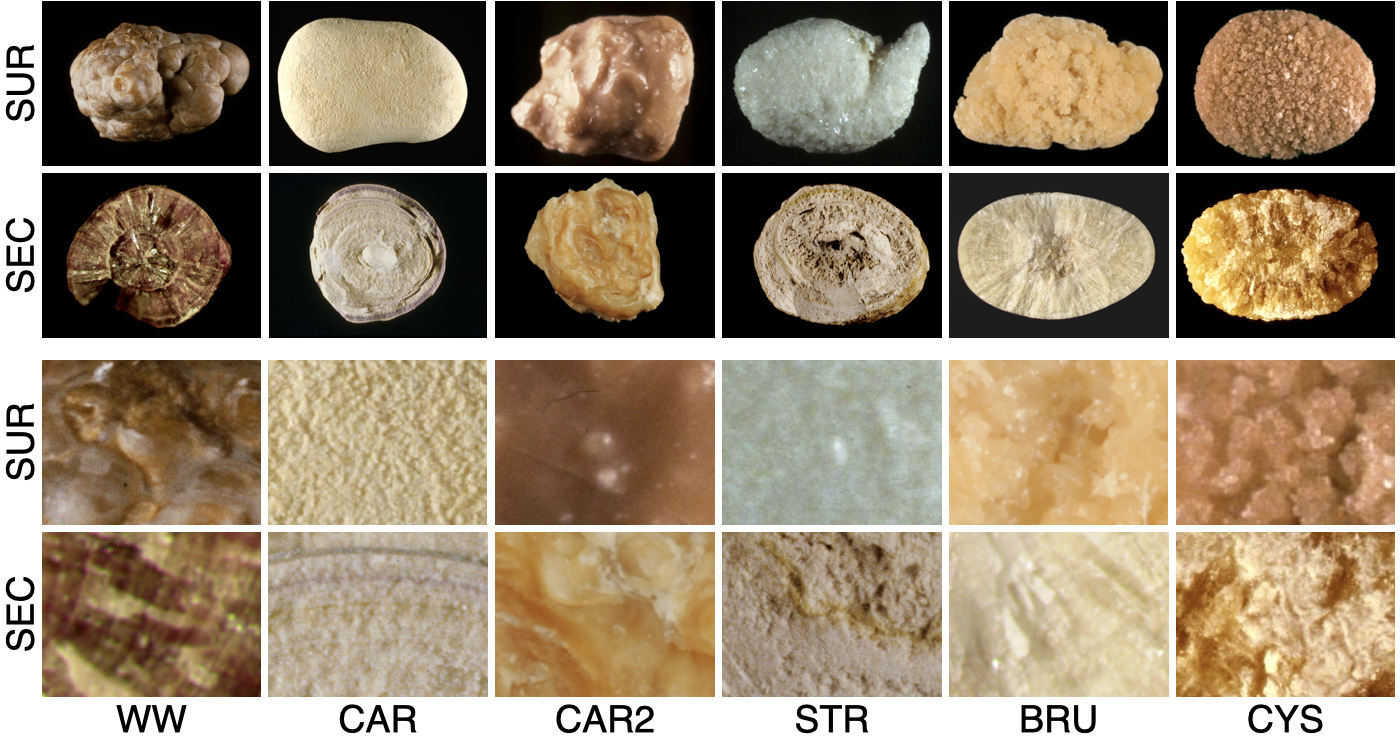}}
    \hspace{1mm}
    \subfloat[Dataset B -- Simulated endoscopy]{
    \label{fig:datasetb}\includegraphics[width=0.485\linewidth]{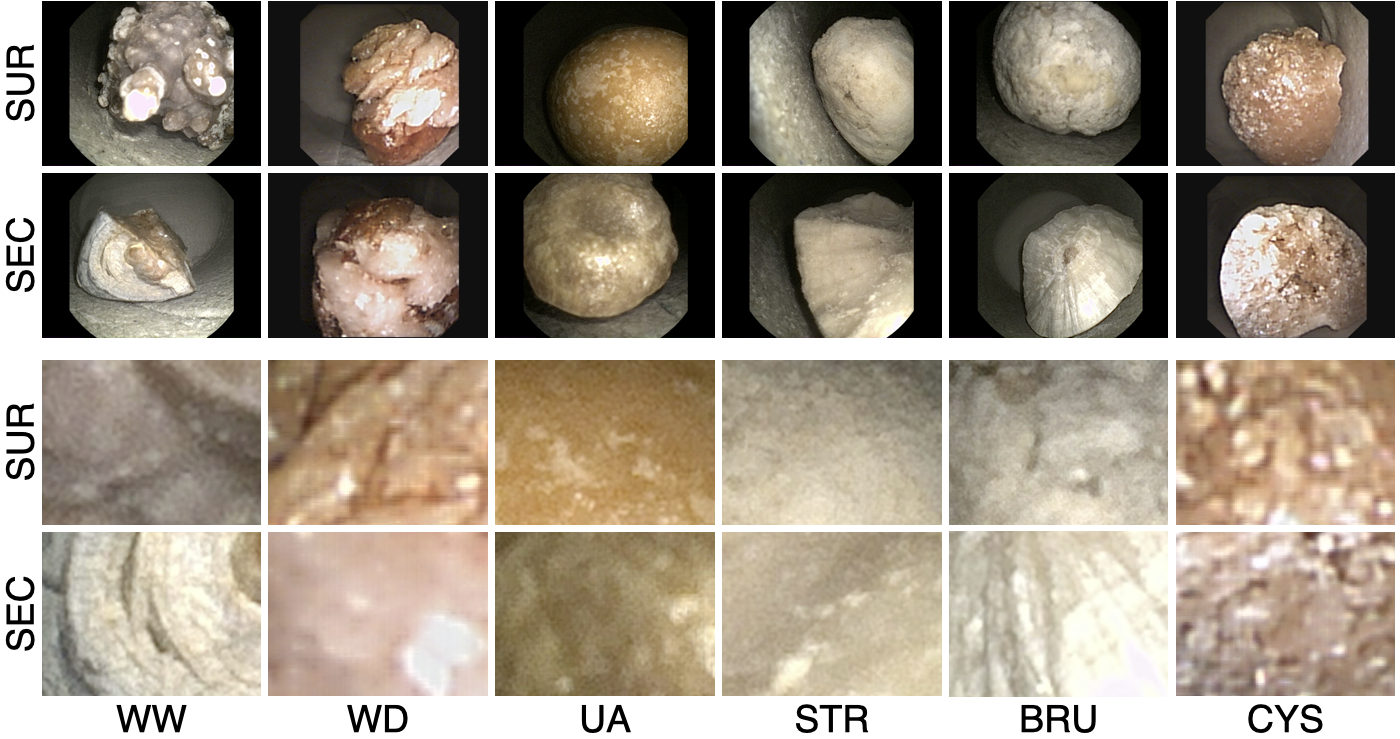}}
    \vspace{-1.5mm}
    \caption{Examples of ex-vivo kidney stone images acquired with two different devices. The two upper lines show complete stone fragment surfaces (SUR) and sections (SEC), whereas the two lower lines give image patches (small fragment parts) extracted from the fragment images. (a) Dataset A acquired with a CCD camera (\cite{corrales2021classification}). The image  background is uniform in color and the six subtypes (WW, CAR, CAR2, STR, BRU and CYS) are  described in section \ref{sec:DataDaudon}. (b) Dataset B acquired with a flexible ureteroscope (\cite{elbeze2022evaluation}). The image background is non uniform and the six subtypes (WW, WD, UA, STR, BRU and CYS) are  described in section \ref{sec:DataElBeze}.    
    % \textcolor{orange}{Faltan imagenes completas de MyStone: Dataset C. <- FOR ME PUT THESE IMAGES and PATCHES OF DATASET C IN NEW FIGURE 5 with the same presentation on 4 lines as in this figure } 
    }
    \label{fig:dataset_private}
\end{figure*}

Three 
% complementary IN what they are complementary ?  
different kidney stone image datasets were employed to assess the performance the proposed Federated Learning framework. 
% we employ  <- Avoid we! IN a paper the style is neutral
These ex-vivo data sets were built in various conditions, i.e., the images were acquired with different sensors and in different environments, and are   
% microscopy imaging ?? workflows
the \textbf{Michel Daudon} dataset (Dataset A, a part of the images described in (\cite{corrales2021classification}), the \textbf{Jonathan El-Beze} dataset (Dataset B, see (\cite{elbeze2022evaluation}), and the \textbf{MyStone} dataset (Dataset C,  (\cite{serrat2017mystone}). Image sets A and B are private datasets (see Fig. \ref{fig:dataset_private}), while  dataset C is a popular and publicly available benchmark. All datasets contain among the six most representative (frequent) kidney stone classes with known ground truth (the subtypes were identified with the reference MCA procedure). The very different acquisition procedures described below allow for an evaluation of the proposed Federated Learning-approach in terms of domain generalization and robustness to acquisition variability.

\subsubsection{Michel Daudon Dataset (Dataset A) \label{sec:DataDaudon}}
The \textit{Michel Daudon} dataset (\cite{corrales2021classification}) is comprised of 366 high-resolution images of kidney stone fragments captured by a CCD (Charge-Coupled Device) camera under a controlled viewpoint. 
%(see Fig. \ref{fig:dataset_private}.(a)). 
Each image with a 4288 $\times$ 2848 pixel resolution was obtained against a uniform background, with a small motion of a handheld camera and with few specular reflections.  
This dataset contains three image type subsets:
\begin{itemize}
    \item \textbf{Section (SEC)} – cross-sectional views of the kidney stone fragments that highlight the internal crystalline composition.
    \item \textbf{Surface (SUR)} – surface views of the renal calculi fragments showing the external stone textures and colors.
    \item \textbf{Mixed (MIX)} – a heterogeneous subset mixing both SEC and SUR views.
\end{itemize}
As illustrated in Fig. \ref{fig:dataset_private}.(a), all three image type subsets of dataset A include six morphological subtypes: Ia-WW (Whewellite, Subtype Ia), IVa-CAR (Carbapatite, Subtype IVa), IVa2-CAR2 (Carbapatite, Subtype IVa2), IVc-STR (Struvite, Subtype IVc),  IVd-BRU (Brushite, Subtype IVd), Va-CYS, (Cystine, Subtype Va). In the two upper lines in Fig. \ref{fig:dataset_private} are images of complete kidney stone fragment surfaces (SUR, first line) and sections (SEC, second line) acquired on a homogeneous background. A total of 18{,}000 image patches (lines 3 and 4 in Fig. \ref{fig:dataset_private}.(a)) were extracted from the 366 high resolution images (the patch extraction method is justified in subsection \ref{sec:patchExtraction})  and used  for training and testing the proposed Federated Learning-approach.  

\begin{comment}
I DO NOT SEE THE INTEREST OF THIS FIGURE. WHAT ADDITIONNAL INFORMATION IS GIVEN ? 
\begin{figure}[h]
\centering
\includegraphics[width=\linewidth]{daudon_samples.png}
\caption{Michel Daudon dataset (Dataset A) showing example \textit{patches} of Carbapatite variants (IVa, IVa2) alongside common stone types, across three imaging modalities (MIX, SEC, and SUR).}
\label{fig:daudoin_samples}
\end{figure}
%
\end{comment}

\subsubsection{Jonathan El-Beze Dataset (Dataset B) \label{sec:DataElBeze}}
The \textit{Jonathan El-Beze} dataset (\cite{elbeze2022evaluation}) consists of 409 endoscopic images of kidney stones captured with a flexible ureteroscope. 
Realistic  acquisition conditions were simulated in the sense that the kidney stone fragments were placed in hollow tubular shaped parts having the dimensions (length and section) and colors (an orange close to that of the inner urinary tract walls) of an ureter. The endoscope was introduced in the simulated hollow organ and the acquisition viewpoint was not always optimal and the illumination  was not diffuse, as during an ureteroscopy and contrary to the conditions of dataset A and C.  The images of this dataset are with a spatial resolution of 1920 $\times$ 1080 pixels and were acquired under inhomogeneous illumination and motion blur.  
Similar to Dataset A, it includes three image subsets:\textit{Section (SEC)}, \textit{Surface (SUR)}, and \textit{Mixed (MIX)}, with six morphological subtypes Ia-WW (Whewellite Subtype Ia), IIa-WD (Weddellite, Subtype IIa), IIIa-UA (Uric Acid Subtype IIIa), IVc-STR (Struvite, Subtype IVc), IVd-BRU (Brushite, Subtype IVd), and Va-CYS (Cystine, Subtype Va). 18{,}000 patches were extracted from the 409 endoscopic images (see Fig. \ref{fig:dataset_private}.(b)).
%This collection introduces valuable inter-laboratory and acquisition variability that reflects the challenges of clinical imaging environments.

\begin{comment}
I DO NOT SEE THE INTEREST OF THIS FIGURE. WHAT ADDITIONNAL INFORMATION IS GIVEN ? 
\begin{figure}[h]
\centering
\includegraphics[width=\linewidth]{jonathan_samples.png}
\caption{Jonathan El-Beze dataset (Dataset B) showing example \textit{patches} of the six kidney stone subtypes across three imaging modalities (MIX, SEC, and SUR).}
\label{fig:jonathan_samples}
\end{figure}
\end{comment}

\subsubsection{MyStone Dataset (Dataset C) \label{sec:MyStoneDataset}}
The \textit{MyStone} dataset (\cite{serrat2017mystone}) is a publicly available collection of ex-vivo kidney stone images
%  microscopy ? NO, certainly not
%images designed to support supervised classification and domain generalization studies. <- Are you sure about generalization?
%
% As I can remember about this publication
%
acquired under ideal conditions, i.e. with both a perfectly controlled CCD-camera viewpoint (the camera is mounted on a support) and illumination source positions. 
This dataset includes a total of 1,426 
%endoscopic <- No !  certainly not again.
images, divided into 1,142 frames for training and 284 other frames for testing. A patch-based data augmentation strategy was used to address class imbalance and increase the number of samples. Specifically, six equally sized classes consist each of 800 training and 200 test patches (i.e., training and test sets of respectively 4,800 and 1,200 patches in all). Unlike the private {\em Michel Daudon} and {\em Jonathan El-Beze} datasets, {\em MyStone} does not distinguish between surface and section views. All patches are extracted from uniformly illuminated images in which the fragments were placed on a background with a constant color.
%
%Each patch is represented in RGB format with a spatial resolution of 255$\times$255 pixels. 
The six classes of the {\em MyStone} dataset
%correspond to specific crystalline compositions, 
include both pure and mixed stone subtypes: Ia-WW (Whewellite, Subtype Ia), IIa-WD (Weddellite, Subtype IIa), HAP (Hydroxyapatite, mixed composition), STR-IVc (Struvite, Subtype IVc), BRU-IVd (Brushite, Subtype IVd), and UA-IIIa (Uric Acid, Subtype IIIa). 

\begin{figure}[!t]
\centering
\includegraphics[width=\linewidth]{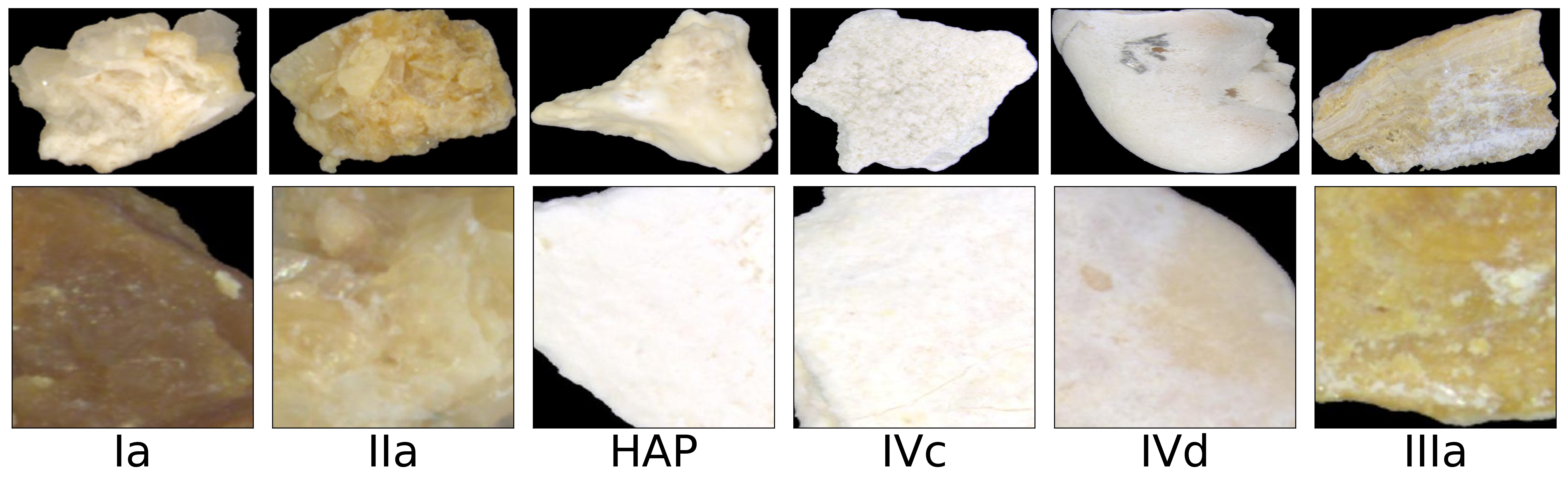}
\caption{Samples of the {\em MyStone} dataset. The first row gives complete stone fragment images (MIX image set type), whereas the second row shows patches extracted from the images (small fragment parts). The image background is uniform in color and the six subtypes (Ia, IIa, HAP, IVc, IVd and IIIa) are described in section \ref{sec:MyStoneDataset}.
}
\label{fig:mystone_samples}
\end{figure}

\begin{table*}[t]
\centering
\caption{Comparison of the content of the three databases in terms of kidney stone subtypes. Each dataset includes six subtypes of the Daudon's worldwide reference classification (\cite{corrales2021classification}).  “Yes” indicates presence of a given subtype in a dataset or accros all datasets for the last column; “Partially” stands for subtypes shared by only two datasets. \label{tab:datasets_comparison}}
\resizebox{\textwidth}{!}{
\begin{tabular}{llcccc}
\toprule
\textbf{Stone Type (morphology)} & \textbf{Subtype} & \textbf{MyStone} & \textbf{Jonathan El-Beze} & \textbf{Michel Daudon} & \textbf{Common Across All} \\
\midrule
Calcium Oxalate Monohydrate (Whewellite) & Ia & Yes & Yes & Yes & Yes \\
Calcium Oxalate Dihydrate (Weddellite) & IIa & Yes & Yes & – & Partially \\
Uric Acid & IIIa & Yes & Yes & – & Partially \\
Hydroxyapatite & HAP & Yes & – & – & No \\
Carbapatite & IVa & – & – & Yes & No \\
Carbapatite Type 2 & IVa2 & – & – & Yes & No \\
Struvite & IVc & Yes & Yes & Yes & Yes \\
Brushite & IVd & Yes & Yes & Yes & Yes \\
Apatite & Va & – & Yes & Yes & Partially \\
%\midrule
%\textbf{Total Images} & --- & 6,000 & 18,000 & 18,000& --- \\
%\textbf{Variants} & --- & 1 (MIX) & 3 (MIX, SEC, SUR) & 3 (MIX, SEC, SUR) & --- \\
%\textbf{Image Format} & --- & JPG (255×255) & PNG (256×256) & PNG (256×256) & --- \\
\bottomrule
\end{tabular}
}
\end{table*}
%
%\vspace*{2mm}
%
\begin{table*}[]
\caption{Overview of the data type subsets (SEC, SUR and/or MIX views) and and the available patches for the three classes.\label{tab:datasets_summary}}
\resizebox{\textwidth}{!}{
\begin{tabular}{lccccc}
\toprule
\textbf{Dataset} & \textbf{Image types} & \textbf{Subtypes/classes} & \textbf{Train patches} & \textbf{Test patches} & \textbf{\textcolor{black}{Total  amount}} \\
\midrule
\textbf{Michel Daudon (A)} & MIX, SEC, SUR & 6 & 14,400 & 3,600 & 18,000 \\
\textbf{Jonathan El-Beze (B)} & MIX, SEC, SUR & 6 & 14,400 & 3,600 & 18,000 \\
\textbf{MyStone (C)} & MIX & 6 & 4,800 & 1,200 & 6,000 \\
\midrule
%\textbf{Image Format} & \multicolumn{5}{l}{A/B: PNG (256×256 RGB); C: JPG (255×255 RGB)} \\
\textbf{Patch Generation} & \multicolumn{5}{l}{Non-overlapping $256\times256$ patches extracted from non-image background regions; prevents data leakage.} \\
\textbf{Split Ratio} & \multicolumn{5}{l}{Data stratification for all datasets:  80\% of patches for training and 20\% for testing.} \\
\bottomrule
\end{tabular}
}
\end{table*}
\subsubsection*{\bf Complementarity of the databases.}
Table \ref{tab:datasets_comparison} highlights the fact that three kidney stone types (subtypes Ia, IVc and IVd) are in all three databases, that three other renal calculi types are in at least two databases (subtypes IIa, IIa and Va) and that three subtypes are only present in one database (HAP, IVa and IVa2).
Table \ref{tab:datasets_summary} gives a general view on the data available in each database: image or view type (only mixed views or also sorted views as SUR or SEC), number of patches, etc.

It is noticeable that the three datasets are different from several viewpoints :  \\
\noindent{-} {\em Acquisition device and image resolution.} Datasets A and C were acquired with different CCD cameras, whereas dataset B was acquired with an endoscope. Both the sensor and its resolution were different for all datasets. \\
\noindent{-} {\em Image quality.} While for dataset C the sensor was mounted on a support, dataset A was acquired with a handheld camera (with small movements) and dataset B with an endoscope involving more movements during the acquisition. The level of motion blur and loss of contrast increases when passing from dataset C to A and from dataset A to B. Dataset B (endoscopy) exhibits also the greatest image illumination inhomogeneity and specular reflection risks, while these effects are respectively less strong and non-existing in datasets A and C. \\
\noindent{-} {\em Viewpoint control.} While the viewpoints were easy to control for datasets A and C, the choice of the camera pose is more difficult to master since an endoscope is less easy to control in a constrained environment that was simulated for dataset B.

\subsubsection{Patch Extraction \label{sec:patchExtraction}}
All three datasets underwent same preprocessing steps. 
% to ensure methodological consistency. <- what does it mean ? 
Each image was manually segmented under expert supervision to isolate the kidney stone regions and exclude surrounding background. Following the protocol described in (\cite{lopez2021assessing}), square patches of size $256 \times 256$ pixels were extracted from the segmented regions to balance the classes by increasing the sample amount for some subtypes. A maximum overlap of 20 pixels was allowed between adjacent patches to minimize redundancy. The optimal patch size of $256 \times 256$ pixels in ureteroscopy was demonstrated in (\cite{lopez2024vivo}).

For the private datasets of {\em Daudon} and {\em El-Beze}, this extraction process yielded approximately 12{,}000 patches per dataset, distributed among the three subsets (SUR, SEC, MIX). Each subset contained 6{,}000 patches (4{,}800 for training and 1{,}200 for testing), ensuring no overlap between sets to prevent data leakage. The MIX subset aggregates both views (SUR and SEC) for a comprehensive representation.

The patches were normalized using a whitening process to reduce the sensitivity to illumination differences. 
%and texture variability. <- no !
In each color channel ($cc$ = $R$, $G$ or $B$) , the means $\mu^{cc}$ and standard deviations $\sigma^{cc}$ were calculated for the patches, and the patch values $P^{cc}(x_i, y_i)$ were normalized using Eq. \eqref{eq:whitening} for each pixel $i$ with coordinates $(x_i, y_i)$.
\begin{equation} \label{eq:whitening}
P_w^{cc}(x_i, y_i) = \frac{P^{cc}(x_i, y_i) - \mu^{cc}}{\sigma^{cc}}
\end{equation}
This color intensity normalization contributes to the cross-domain generalization.
%produced consistent pixel value distributions <- MEANS NOTHING!  
%across all datasets, 

\subsubsection{Dataset Partitioning and CNN-Input Adaptation.}
\label{subsection:DataPartandProcessing}
All datasets were partitioned using a stratified and non-overlapping 80\%/20\% split between training and testing subsets to guarantee a balanced representation of all kidney stone subtypes. 
No spatial overlap or duplication of patches was allowed across partitions to avoid data leakage (i.e., in each partition a patch of a given image region appears only once). 
%each patch was derived from a unique image region. <-

%After extraction, all patches were uniformly standardized to a spatial resolution of \(256\times256\) pixels with three RGB channels. <- ALREADY DONE SINCE THE PATCHES HAVE THIS SIZE! 

Pixel intensities were normalized using ImageNet preprocessing statistics (i.e., for the means $[\mu^R, \mu^G,\mu^B]$ = [0.485, 0.456, 0.406] and $[\sigma^R, \sigma^G,\sigma^B]$ = [0.229, 0.224, 0.225] for the standard deviations, ensuring feature-scale consistency with pretrained models.  
This preprocessing step facilitates the direct integration of the datasets into deep convolutional architectures (e.g., ResNet-18), while preserving comparability across the heterogeneous imaging sources used in our federated learning framework.

In general, this common preprocessing and partitioning strategy establishes a consistent experimental foundation across all domains of datasets A, B and C, ensuring fair and reproducible evaluations within the federated learning setting. It also enables direct assessment of cross-domain generalization, a critical aspect for developing robust models capable of handling real-world clinical variability.

% \begin{figure*}[!bt]
% \centering
% \includegraphics[width=\linewidth]{cifar10_mnist_combined.png}
% \caption{Sample images from the ten object classes in the CIFAR-10 dataset (upper line) and the ten digits in MNIST dataset (lower line). \textcolor{blue}{IS THIS FIGURE REALLY OF  INTEREST ? It takes place for known pictures and the paper is already long. Description in text not sufficient?} \textcolor{red}{Too much space betwwen the lines of pictures.}}
% \label{fig:cifar10_mnist_samples}
% \end{figure*}

\subsection{Benchmark Validation on MNIST and CIFAR-10.}
%, we validate its  AVOIS we, us ... 
The performance of the proposed FedAgain framework was validated using two widely adopted public benchmarks to demonstrate its general applicability  beyond medical imaging.
These benchmarks are the MNIST and CIFAR‑10 datasets. These datasets allow the evaluation of the proposed method on both grayscale and RGB inputs under well-studied conditions, providing comparability with the broader state of the art.

The MNIST dataset comprises 60{,}000 training images and 10{,}000 test images of handwritten digits, each in grayscale format with size \(28\times28\) pixels (\cite{deng2012mnist}). Its simplicity and long history in machine-learning research make it an excellent baseline for assessing algorithmic behavior on monochromatic data. On the other hand, the CIFAR-10 dataset consists of 60{,}000 color images of size \(32\times32\), divided into ten object classes (e.g., airplane, car, cat, dog), of which 50{,}000 are used for training and 10{,}000 for testing (\cite{Krizhevsky09learningmultiple}). This dataset introduces the additional complexity of RGB colour channels and a more challenging object-recognition problem.

The application of FedAgain to both datasets allows to:
\begin{itemize}
  \item Evaluate the robustness of trust-weighted aggregation in standard federated setups over heterogeneous clients.
  \item Compare performance on grayscale inputs (MNIST) versus RGB inputs (CIFAR-10) to assess generalization across modalities.
  \item Provide a benchmark against existing literature so that the improvements achieved with FedAgain can be placed in the context of the broader machine-learning community.
\end{itemize}

In the proposed experiments, the federated data partitioning mimics non-IID conditions among clients, and FedAgain is applied to aggregate local updates under this heterogeneity. The results on MNIST and CIFAR-10 serve both as sanity checks and as proof-of-concept that FedAgain is versatile and effective beyond the specialized medical domain.
  
\section{Experimental Setup and Evaluation. \label{Sec:Experimental}}
\subsection{Federated Learning Configuration. \label{subsection:FLconfig}}
The federated experiments were conducted using a custom PyTorch-based framework specifically designed for medical imaging applications. This lightweight implementation avoids dependencies on existing libraries such as Flower or TensorFlow Federated, providing full transparency and flexibility to integrate custom aggregation rules. The system comprises a central \textit{Federated Server} coordinating multiple \textit{Federated Clients}, each representing an independent medical site performing local training and evaluation.
\subsubsection{Clients and Communication.} The number of participating clients was configurable between 10 and 20, with each client holding a private dataset that cannot be shared. Communication followed a synchronous, round-based protocol: the central server \emph{ResNet18} model was broadcast to all clients at the start of each round; clients performed local updates for one local epoch (\(E=1\)), and returned model parameters and metrics to the server. The server then aggregated updates according to the selected strategy and redistributed the new global model for the next round. Each experiment used \(R=10\) global rounds, a batch size of 32, and a local learning rate of 0.01.

\subsubsection{Compared Aggregation Strategies.}
The proposed method is compared against several standard and robust aggregation strategies commonly used in federated learning. The formal definitions of these algorithms are provided in Section \ref{sec:preliminaries}. The evaluated aggregation strategies include \textbf{FedAvg} (\cite{mcmahan2017communication}), \textbf{FedProx} (\cite{li2020federated}), \textbf{FedMedian} (\cite{yin2018byzantine}), \textbf{Bulyan} (\cite{liu2023byzantine}), and the proposed method \textbf{FedAgain} (see Section \ref{sec:fedagain}).

\subsubsection{Data Distributions (IID vs Non-IID).} Two partitioning schemes were used:
\begin{enumerate}
    \item \textbf{Independent and Identically Distributed (IID)}  stan\-ds for data that are uniformly distributed across all clients. IID-data serve to establish a baseline for convergence speed.
    \item \textbf{Non-IID} refers to heterogeneous partitions reflecting realistic clinical variability. Two methods were employed: i) \textit{label-skew}, where each client specializes in two morphological classes and ii) \textit{Dirichlet-based sampling} with concentration parameter \(\alpha= 0.5\), controlling the degree of heterogeneity.
\end{enumerate}

\subsubsection{Corrupted-Client Simulation.}  
%A fraction \(\rho \in \{0,\,0.1,\,0.3,\,0.5\}\) of clients were artificially corrupted To evaluate the robustness of the Federated Learning-strategies. 
% \textcolor{blue}{the meaning of $\rho$ is really unclear!}
A percentage $\rho_k$ of training images of client $k$ is corrupted by controlled defaults. $\rho_k$ is randomly chosen in $[0\%, 10\%, 30\%, 50\%]$ for each client $k$. 
The images were corrupted by a suite of standardized distortions as done by the common corruption benchmark introduced by (\cite{hendrycks2019benchmarking}). In this setup, the training images of the selected clients are modified by adding them noise, blur, contrast variations, etc. These image modifications simulate complex acquisition conditions or device heterogeneity in federated settings.
%
% \textcolor{blue}{SE HABLA DE RESULTADOS HASTA SECCIONDE RESULTADOS: 
% These corruptions affected only training data, allowing evaluation on clean test sets to quantify degradation transfer. FedAgain consistently demonstrated superior tolerance to corrupted or low-quality clients.}

Overall, this way to proceed leads to a reproducible and realistic experimental setup that captures both data heterogeneity and quality variability inherent to multi-site federated medical imaging.
\subsection{Dataset Distribution Across Clients \label{subsec:data_distribution}}
\subsubsection{Data Partitioning Strategy}  
Three different partitioning strategies were employed to simulate realistic federated‐learning scenarios where data is naturally heterogeneous across participants: IID, Non‐IID with label skew, and Non‐IID with Dirichlet distribution. In the IID setting, training data is randomly and uniformly partitioned among \(|\mathcal{K}|\) clients (or participating nodes), ensuring that each client receives a balanced representation of all classes. Formally, one have for all clients $k \in [1, \ldots, |\mathcal{K}|]$: 
\begin{equation} \label{eq:IIDpartition}
\mathcal{TD}_k = \{(x_i,y_i)\}_{i \in \mathcal{IS}_k} \;\; \textrm{and} \;\; |\mathcal{IS}_k| \approx \frac{|\mathcal{TD}_{\mathrm{tot}}|}{|\mathcal{K}|},
\end{equation}
where:  
\begin{itemize}
 % \item \(N\) denotes the total number of clients (or participating nodes) in the federated system.
  \item $x_i$ and $y_i$ stand for a training image and its class label (ground truth), respectively.  
  \item \(\mathcal{TD}_{tot}\) represents the entire training dataset of \(|\mathcal{TD}_{tot}|\) images 
  %centrally aggregated  <- THIS IS CONFUSING, to be removed
  before partitioning (see Eq. \eqref{eq:1}).
  \item \(\mathcal{TD}_k\) is the local training image set assigned to client \(k\) and \(|\mathcal{TD}_k|\) the amount of these images (see Eq. \eqref{eq:1}).
  \item $\mathcal{IS}_k \subset \{1,\dots,|\mathcal{D}_{\mathrm{tot}}|\}$ is the index set of the part of training images of $\mathcal{TD}_{tot}$ allocated to client \(k\). The amount of entries $|\mathcal{IS}_k|$ in index set $\mathcal{IS}_k$ is equal to  $|\mathcal{TD}_k|$. The approximation in Eq. \eqref{eq:IIDpartition} is due to the fact that, in IID scenarios, the images (here of kidney stone subtypes) cannot be completely uniformly distributed over the same classes located on the different clients.  
\end{itemize}
This baseline scenario is classically taken as reference, although almost uniform distributions rarely occur in real‐world federated deployments, where participants (for example, hospitals, mobile devices, or edge sensors) typically possess data with distinct local characteristics.

\subsubsection{Non-IID Label Skew Partitioning}  
A label-skew strategy was implemented to conduct the controlled non-IID experiments. In such a strategy, each client \(k\) specializes in a subset of classes while maintaining limited exposure to others. Specifically, each client is assigned with a given amount of primary classes 
%(default: \(C_{\text{primary}}=2\)), SHOULD WE REALLY SAY THIS HERE ? 
%from which it receives 
representing a dominant portion (i.e., 80\% in Eq. \eqref{eq:nonIIDpartition}) of the local data set, while the remaining classes contribute equally to the other 20\% of the local learning samples to prevent complete specialization. Formally, 
%for client \(k\) 
for class $c$ which is a primary class on $|\mathcal{K}_\textrm{primary}^{\, c}|$ clients $k$ included in set $\mathcal{K}_\textrm{primary}^{\, c}$, 
%\(\mathcal{C}_k \subset \{c_1,c_2,\dots,c_{|\mathcal{K}|}\}\), 
the number of samples $n_{k,c}$ of class \(c\) on all clients $k$ is given by:  
\begin{equation} \label{eq:nonIIDpartition}
n_{k,c} = 
\begin{cases}
 0.8 \cdot \frac{|\mathcal{TD}_{\text{tot}}^{(c)}|}{|\mathcal{K}_\textrm{primary}^{\, c}|} & \text{if } k \in \mathcal{K}_\textrm{primary}^{\, c},\\
 0.2 \cdot \frac{|\mathcal{TD}_{\text{tot}}^{(c)}|}{ |\mathcal{K}| - |\mathcal{K}_\textrm{primary}^{\, c}|} & \text{otherwise},
\end{cases}
\end{equation}
where \(\mathcal{TD}_{\text{tot}}^{(c)}\) denotes the total amount of training samples of class \(c\), and $|\mathcal{K}| > |\mathcal{K}_\textrm{primary}^{\, c}|$ remains the total number of clients in set \(\mathcal{K}\). Depending on the values of  $|\mathcal{K}|$, $|\mathcal{K}_\textrm{primary}^{\, c}|$  and $\mathcal{TD}_{\text{tot}}^{(c)}$, $n_{k,c}$ can be a real value that must be rounded to the nearest integer.

This approach creates interpretable and reproducible heterogeneity patterns, facilitating a systematic analysis of the strategy performance under predictable non-IID conditions.
\subsubsection{Dirichlet Distribution Partitioning}  
The widely-used Dirichlet distribution partitioning me\-thod was adopted to simulate realistic and stochastic forms of client heterogeneity. This method has become a standard benchmark in federated learning research (\cite{hsu2019nonIID}).  

For each class \(c \in [1, 2, \dots, |\mathcal{C}|]\),  a probability vector $\mathbf{p}_c$ is sampled from a symmetric Dirichlet distribution with concentration parameter \(\alpha\). 
\begin{equation} \label{eq:DDP}
\mathbf{p}_c = (p_{c,1}, p_{c,2}, \dots, p_{c,|\mathcal{K}|}) \;\sim\; \mathrm{Dirichlet}(\alpha, \alpha, \dots, \alpha)
\end{equation} 
 In Eq. \eqref{eq:DDP}, \(p_{c,k}\) denotes the fraction of class-\(c\) samples that are assigned to client \(k \in \mathcal{K}\).  

Once \(\mathbf{p}_c\) is drawn, the total number of samples of class \(c\) in the training set, denoted by \(\lvert \mathcal{D}_{\mathrm{tot}}^{(c)} \rvert\), is distributed to each client \(k\) by respecting class sample sizes given by $p_{c,k} \times \mathcal{D}_{\mathrm{tot}}^{(c)}$.

The concentration parameter \(\alpha\) controls the degree of non‐IIDness: when the \(\alpha\)-value tends to 0, the partitioning becomes highly imbalanced (i.e., clients tend to specialize in very few classes), whereas as for $\alpha$ tending to $+\infty$, the distribution converges toward uniform (IID) across clients. Following established conventions in Federated Learning literature (\cite{reguieg2023comparative}),  \(\alpha\) was set at 0.5 in the proposed experiments to generate challenging non‐IID partitions that reflect realistic multi‐site scenarios, while preserving sufficient data diversity for convergence.  

This partitioning method produces naturally varying label proportions across clients, allowing the evaluation to rigorously test the robustness of model aggregation methods in federated settings characterized by unpredictable and heterogeneous data distributions.
\begin{table*}[h]
\centering
\caption{Summary of corruption types used to simulate image degradation. Corruptions are grouped into four categories: noise, blur, weather, and digital distortions, reflecting common sources of quality degradation encountered in real-world imaging benchmarks and medical scenarios.}
\label{tab:corruption_types}
\begin{tabular}{lllp{5cm}}
\toprule
\textbf{Category} & \textbf{Corruption Type} & \textbf{Description} & \textbf{Real-World Cause} \\
\midrule
\multirow{4}{*}{\textbf{Noise}} 
& Gaussian Noise & Additive white Gaussian noise & Sensor thermal noise in imaging hardware \\
& Shot Noise & Poisson-distributed photon noise & Low-light or high-sensitivity imaging \\
& Impulse Noise & Salt-and-pepper pixel dropouts & Transmission or sensor readout errors \\
& Speckle Noise & Multiplicative noise pattern & Ultrasound or coherent imaging interference \\
\midrule
\multirow{5}{*}{\textbf{Blur}} 
& Defocus Blur & Out-of-focus blurring & Lens misalignment or depth-of-field limitation \\
& Glass Blur & Diffusion through translucent medium & Imaging through barriers (e.g., glass) \\
& Motion Blur & Directional streaking & Camera or patient motion during capture \\
& Zoom Blur & Radial stretching artifacts & Optical zoom during exposure \\
& Gaussian Blur & Uniform isotropic smoothing & Lens imperfections or post-processing filters \\
\midrule
\multirow{3}{*}{\textbf{Weather}} 
& Snow & White particle overlays & Adverse atmospheric or environmental particles \\
& Frost & Ice crystal overlay & Low-temperature conditions on optics or coverings \\
& Fog & Haze and reduced contrast & Scattering and reduced visibility in imaging \\
\midrule
\multirow{7}{*}{\textbf{Digital}} 
& Brightness & Global intensity shift & Over/underexposure during acquisition \\
& Contrast & Contrast gain modification & Uneven illumination or sensor calibration \\
& Elastic Transform & Non-rigid geometric deformation & Lens distortion or localized warping \\
& Pixelate & Coarse pixel downsampling & Limited sensor resolution or compression \\
& JPEG Compression & Block artifacts and ringing effects & High compression rates in storage or transmission \\
& Spatter & Overlay of stain-like textures & Contamination on optical elements or lens \\
& Saturate & Altered color intensity & White balance or color calibration errors \\
\bottomrule
\end{tabular}
\end{table*}

\subsection{Corruption Simulation and Robustness Tests}

Corrupted clients representing medical centers affected by degraded data quality were simulated to evaluate robustness under realistic conditions. 
Corruptions included noise, blur, and illumination artifacts, mimicking sensor malfunction, motion blur, or compression errors in medical imaging (see Table \ref{tab:corruption_types}). This approach follows the CIFAR-10-C and ImageNet-C protocols (\cite{hendrycks2019benchmarking}), adapted to the classification of kidney stones.

\noindent\textbf{Corruption Ratio (\(\rho\)).} The corruption ratio determines the fraction of clients receiving corrupted data during training. Four settings were evaluated, \(\rho \in \{0, 0.1, 0.3, 0.5\}\), representing clean, low, moderate, and highly degraded conditions, respectively. In the default configuration, \(\rho = 0.3\) (30\% of clients) was used to model a realistic Byzantine scenario. Corruption was applied only to the training data of affected clients, while the global model was always evaluated on clean test sets to measure generalization to uncontaminated data.

% \noindent\textbf{Corruption Severity.} Each corrupted client applied one randomly selected corruption per image using the image corruptions proposed by  (\cite{hendrycks2019benchmarking}). Severity levels $s$ ranged from 1 (minimal) to 5 (extreme), controlling the intensity of the degradation. The default configuration used a medium level (\(s=3\)), corresponding to noticeable but interpretable visual distortions. The corruption types spanned 16 categories across four families: \textit{noise} (e.g., Gaussian, shot, impulse, speckle), \textit{blur} (defocus, motion, glass, zoom), \textit{weather} (fog, frost, snow) \textcolor{blue}{<- interest in medical application?}, and \textit{digital} (contrast, pixelation, compression).

\noindent\textbf{Corruption Severity.} Each corrupted client applied one randomly selected corruption per image using the image corruptions proposed by \cite{hendrycks2019benchmarking}. Severity levels $s$ ranged from 1 (minimal) to 5 (extreme), controlling the intensity of degradation. The default configuration used a medium level ($s=3$), corresponding to noticeable but diagnostically interpretable visual distortions. The corruption types spanned 16 categories across four families: \textit{noise} (e.g., Gaussian, shot, impulse, speckle), \textit{blur} (defocus, motion, glass, zoom), \textit{weather} (fog, frost, snow), which serve as proxies for acquisition and environmental artifacts encountered in real clinical imaging scenarios, and \textit{digital} (contrast, pixelation, compression).

\begin{figure*}[h]
\centering
\includegraphics[width=\linewidth]{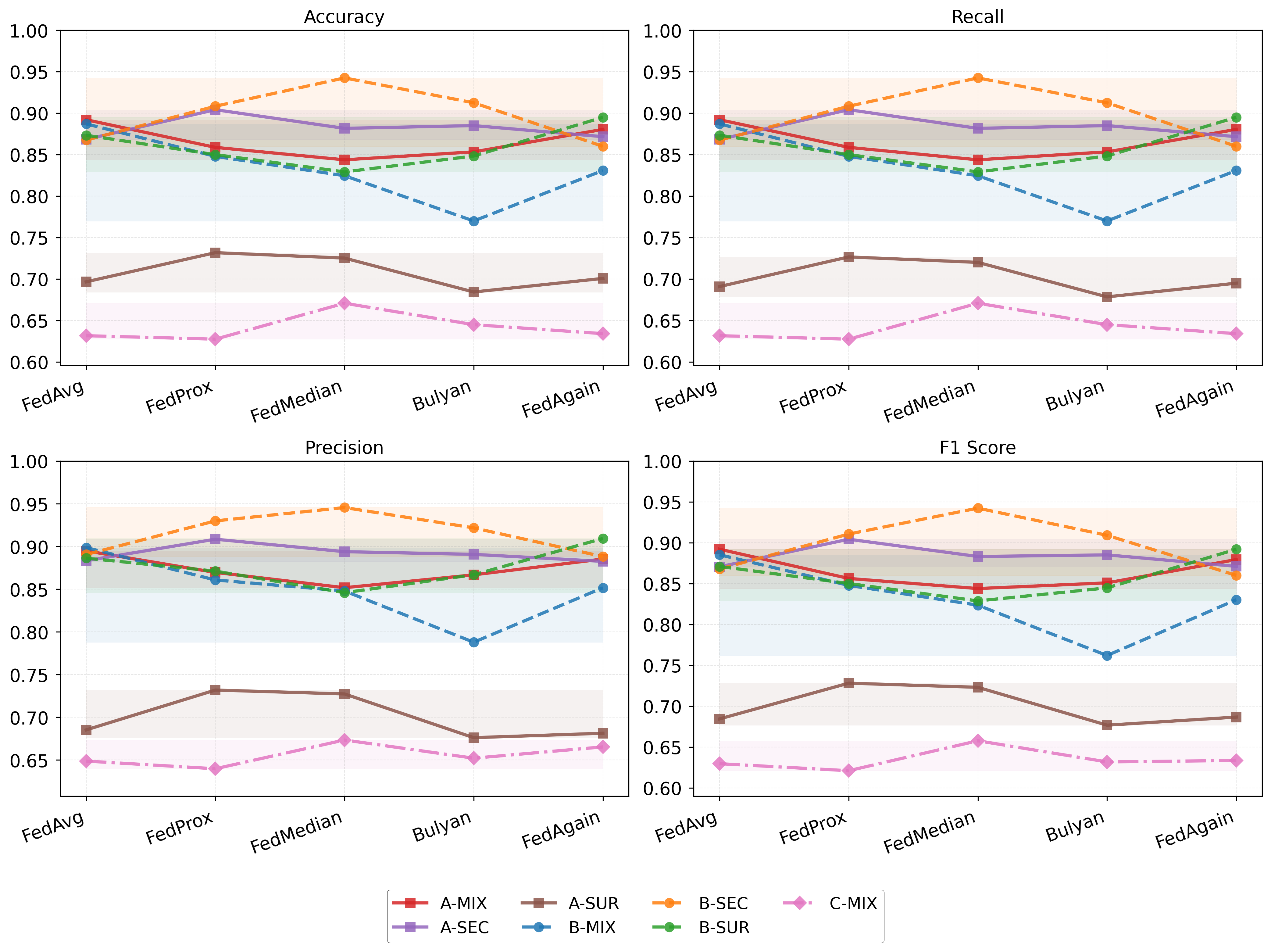}
\caption{Performance comparison of federated learning strategies on kidney stone datasets under clean conditions (IID distribution, 10 clients, 1 round). The evaluated datasets include A (Daudon), B (Jonathan), and C (MyStone). Shaded bands represent the performance range of each dataset across all five aggregation strategies. The nearly horizontal curves and narrow bands indicate that, in the absence of corruption, the choice of aggregation strategy has minimal impact on model performance. Instead, the observed variations are primarily driven by the intrinsic complexity of the datasets rather than by the aggregation method itself.}
\label{fig:kidney_stone_no_corruption}
\end{figure*}

\noindent\textbf{Experimental Design.} For each aggregation strategy, experiments were performed across all corruption ratios and severity levels, forming a robustness matrix:
\begin{itemize}
    \item \(\rho = 0\): baseline, clean training.
    \item \(\rho = 0.1, 0.3, 0.5\): increasing corruption ratios.
    \item \(s \in \{1,2,3,4,5\}\): severity sweep (FedAvg vs. FedAgain comparison).
\end{itemize}
This produced more than 90 experimental configurations across the three datasets. Performance degradation under corruption was used to quantify robustness. 
In general, this corruption simulation protocol provides a rigorous and reproducible benchmark to assess the robustness of federated learning to heterogeneity of data-quality in real-world medical imaging scenarios.

\section{Experimental Results and Discussion}

\subsection{Baseline Evaluation under Clean IID Conditions}

The performance of all federated learning strategies was first evaluated under ideal conditions without data corruption
to establish a baseline for comparison. This experiment was conducted using an IID data partitioning scheme across 10 clients, trained for 10 rounds. This configuration allows to assess the inherent characteristics of each aggregation strategy in the absence of adversarial conditions.

Figure \ref{fig:kidney_stone_no_corruption} presents the comparative performance of five federated learning strategies: FedAvg, FedProx, FedMedian, Bulyan and FedAgain (the proposed one) are evaluated on seven datasets of kidney stones using four key metrics: \emph{Accuracy}, \emph{Recall}, \emph{Precision} and \emph{F1 Score}. The datasets include three subsets of Dataset A (Daudon: A-MIX, A-SEC, A-SUR), three subsets of Dataset B (Jonathan: B-MIX, B-SEC, B-SUR) and Dataset C (MyStone: C-MIX).

Under clean data conditions, all strategies demonstrated relatively similar performance in most metrics, as expected when no malicious clients were present. For datasets (A-SEC, A-MIX, B-SEC, and B-SUR) that led to a high performance, the accuracy values ranged from approximately 84\% to 95\% for FedProx which also slightly outperforms other strategies in Dataset B-SEC (94.8\% accuracy), followed closely by A-SEC (90.4\% accuracy).

The traditional FedAvg baseline achieved competitive results, with accuracy values between 86.8\% and 89.2\% for the best performing datasets, demonstrating that in clean IID settings, the added complexity of Byzantine-robust aggregation methods (FedMedian, Bulyan) or performance-aware strategies (FedAgain) provides minimal advantage.

These results align with theoretical expectations: in the absence of corruptions or statistical heterogeneity, gradient averaging methods converge to similar solutions, as all client updates contribute constructively toward the global optimum. The observed baseline performance therefore serves as a reference point for evaluating the robustness of each strategy when Byzantine clients are introduced in later experiments.

\subsection{Performance Analysis Under Various Corruption Scenarios}

%
%%%%%%%%%%%%%%%%%%%%%  CONTINUER ICI
%

\begin{figure}[h]
\centering
\includegraphics[width=\columnwidth]{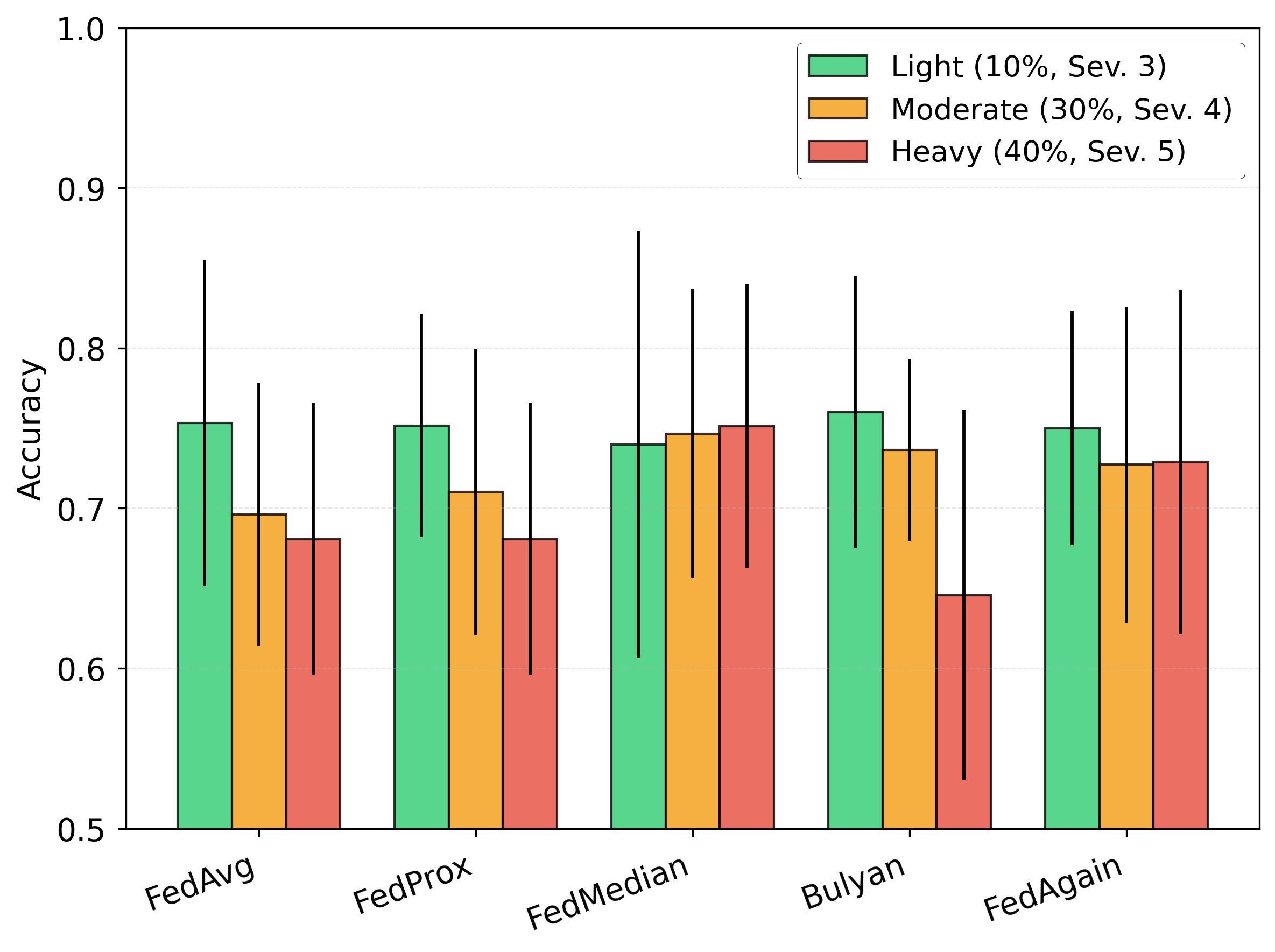}
\caption{Performance comparison of federated learning strategies under light (10\%), moderate (30\%), and heavy (40\%) corruption levels across the seven kidney stone datasets. Bars indicate mean scores with the associated standard deviations.}

% \textcolor{blue}{1) The text over the figure is too small and can be removed (it brings no additional information in comparison to the caption).\newline 2) Only accuracy is discussed in Section 6.2 (nothing about Recall, Precision and F1 score)}

\label{fig:corruption_levels_comparison}
\end{figure}

Experiments under three distinct corruption scenarios were conducted to evaluate the robustness of federated learning strategies against adversarial conditions. The three corruption levels are \emph{``Light''} (10\% of corrupted clients, severity 3), \emph{``Moderate''} (30\% of corrupted clients, severity 4), and \emph{``Heavy''} (40\% of corrupted clients, severity 5). All expe\-ri\-ments were performed with the same configuration as that of the baseline (i.e., 10 clients, 10 rounds, and IID data distribution), with the difference that the corrupted clients were Byzantine clients that generated corrupted gradient updates to simulate adversarial behavior or system failures.

Figure \ref{fig:corruption_levels_comparison} presents the comparative performance of the five federated learning strategies at the three corruption levels, averaged in the seven data sets on kidney stone classification. The error bars represent the standard deviation between different datasets and experimental runs, indicating the variability in the performance of the strategy under adversarial conditions.

Under light corruption conditions (10\% of corrupted clients), all strategies demonstrated relatively robust performance, with accuracy ranging from 73.98\% (FedMedian) to 75.99\% (Bulyan). The minimal performance gap between strategies (approximately 2\%) suggests that when corruption is limited, both standard averaging methods (FedAvg and FedProx) and Byzantine-robust aggregation schemes (FedMedian, Bulyan and FedAgain) can effectively maintain model quality. FedAvg achieved 75.31\% accuracy, performing comparably better to more advanced aggregation methods, indicating that mild corruption can be partially absorbed through simple averaging.

As corruption intensity increased to moderate levels (30\% of corrupted clients, severity 4), a clearer separation in robustness emerged among strategies. FedAgain maintained stable performance with 72.72\% accuracy, exhibiting only marginal degradation compared to the light corruption setting. While Bulyan achieved a comparable accuracy (73.64\%) and FedAvg obtained 69.62\%, FedAgain demonstrated lower performance variability across runs, indicating greater robustness under moderate corruption.

Although FedMedian reached a slightly higher mean accuracy (74.66\%), FedAgain distinguished itself by sustaining nearly constant performance across corruption levels, confirming the benefit of its dual-signal trust weighting mechanism in mitigating the influence of Byzantine clients.

The heavy corruption scenario (40\% of corrupted clients, severity 5) further highlighted FedAgain's resilience. Despite the increased adversarial pressure, it maintained 72.89\% accuracy virtually unchanged from the moderate scenario demonstrating remarkable robustness and stability where most other strategies deteriorated. In contrast, FedMedian showed modest improvement to 75.12\%, while Bulyan degraded sharply to 64.58\% and FedAvg continued to decline to 68.07\%. The consistent performance observed across corruption levels suggests that FedAgain’s adaptive trust-based aggregation effectively filters unreliable client updates and promotes stable convergence, providing a robust solution under increasingly adversarial conditions.

%stabilizes global  stabilizes means nothing, why global ?  

\begin{figure*}[h]
\centering
\includegraphics[width=\linewidth]{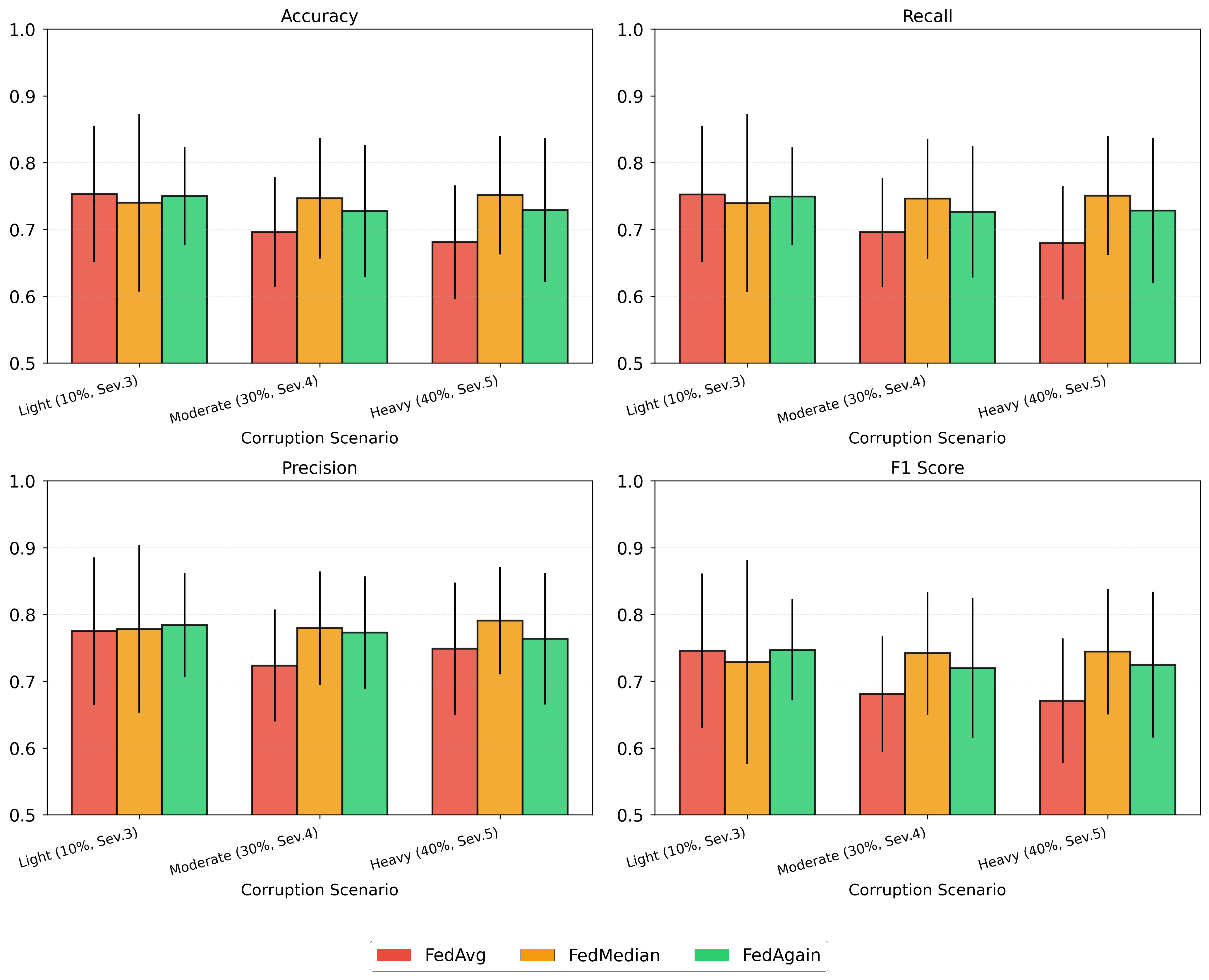}
\caption{Direct comparison of FedAgain, FedAvg, and FedMedian across three corruption scenarios under matched conditions (non-IID label distribution). Each scenario combines a corruption ratio with its corresponding severity level: Light (10\%, Sev. 3), Moderate (30\%, Sev. 4), and Heavy (40\%, Sev. 5). FedAgain consistently outperforms FedAvg across all scenarios and evaluation metrics, with improvements reaching up to +7.1\% in accuracy under heavy corruption. Notably, FedMedian achieves comparable or slightly higher accuracy than FedAgain under light and heavy corruption. However, FedAgain remains competitive while providing the additional advantage of not requiring prior knowledge of the number of Byzantine clients.}
\label{fig:fedagain_vs_fedavg_severity}
\end{figure*}

\subsection{Adaptive Robustness Under Different Corruption Scenarios}

A systematic comparison across three corruption scenarios was conducted to evaluate the adaptive robustness of FedAgain relative to FedAvg and FedMedian under a non-IID label distribution: Light (10\%, Severity 3), Moderate (30\%, Severity 4), and Heavy (40\%, Severity 5). Figure~\ref{fig:fedagain_vs_fedavg_severity} presents the performance of all three strategies across accuracy, recall, precision, and F1-score, averaged over seven sub-datasets for kidney stone classification.

Under Light corruption, all three methods achieved comparable accuracy ($\sim$75\%) and recall ($\sim$74--75\%). However, FedAvg precision dropped to approximately 67\%, while FedMedian ($\sim$78\%) and FedAgain ($\sim$79\%) maintained substantially higher values, representing a gain of roughly 12 percentage points for FedAgain over FedAvg, indicating that FedAvg is already prone to false positives even under mild adversarial conditions.

As corruption intensified to the Moderate scenario, FedAvg accuracy declined to approximately 70\%, while FedMedian and FedAgain maintained values around 74\% and 73\%, respectively. FedAgain matched or slightly exceeded FedMedian in recall ($\sim$75\% vs. $\sim$74\%), with both methods maintaining precision near 78\% compared to FedAvg's 72\%. In F1-score, FedMedian led slightly ($\sim$75\%) over FedAgain ($\sim$73\%), with FedAvg trailing at approximately 68\%.

Under Heavy corruption, FedAvg suffered the most pronounced degradation, with accuracy and recall dropping to approximately 68\%. FedMedian demonstrated resilience in accuracy ($\sim$75\%) and precision ($\sim$79\%), while FedAgain remained competitive at approximately 73\% accuracy and 76\% precision. In F1-score, FedMedian achieved $\sim$75\%, followed by FedAgain ($\sim$71\%) and FedAvg ($\sim$67\%).

An important observation from the figure is the behavior of variance: FedAvg exhibited noticeably larger error bars, especially in precision under light corruption and accuracy under heavy corruption, while FedAgain maintained relatively compact error bars across all metrics, suggesting more stable and predictable behavior across heterogeneous datasets.

Although FedMedian achieved comparable or marginally superior performance in several metrics, this advantage requires prior knowledge of the number of Byzantine clients, information rarely available in practice. FedAgain operates without such assumptions while delivering consistent improvements over FedAvg of approximately 3--7\% in accuracy and 4--12\% in precision across all scenarios, making it particularly suitable for safety-critical medical imaging applications where both diagnostic reliability and practical deployability are essential.

\subsection{Performance Analysis Across Client Data Distributions}

\begin{figure*}[h]
\centering
\includegraphics[width=\linewidth]{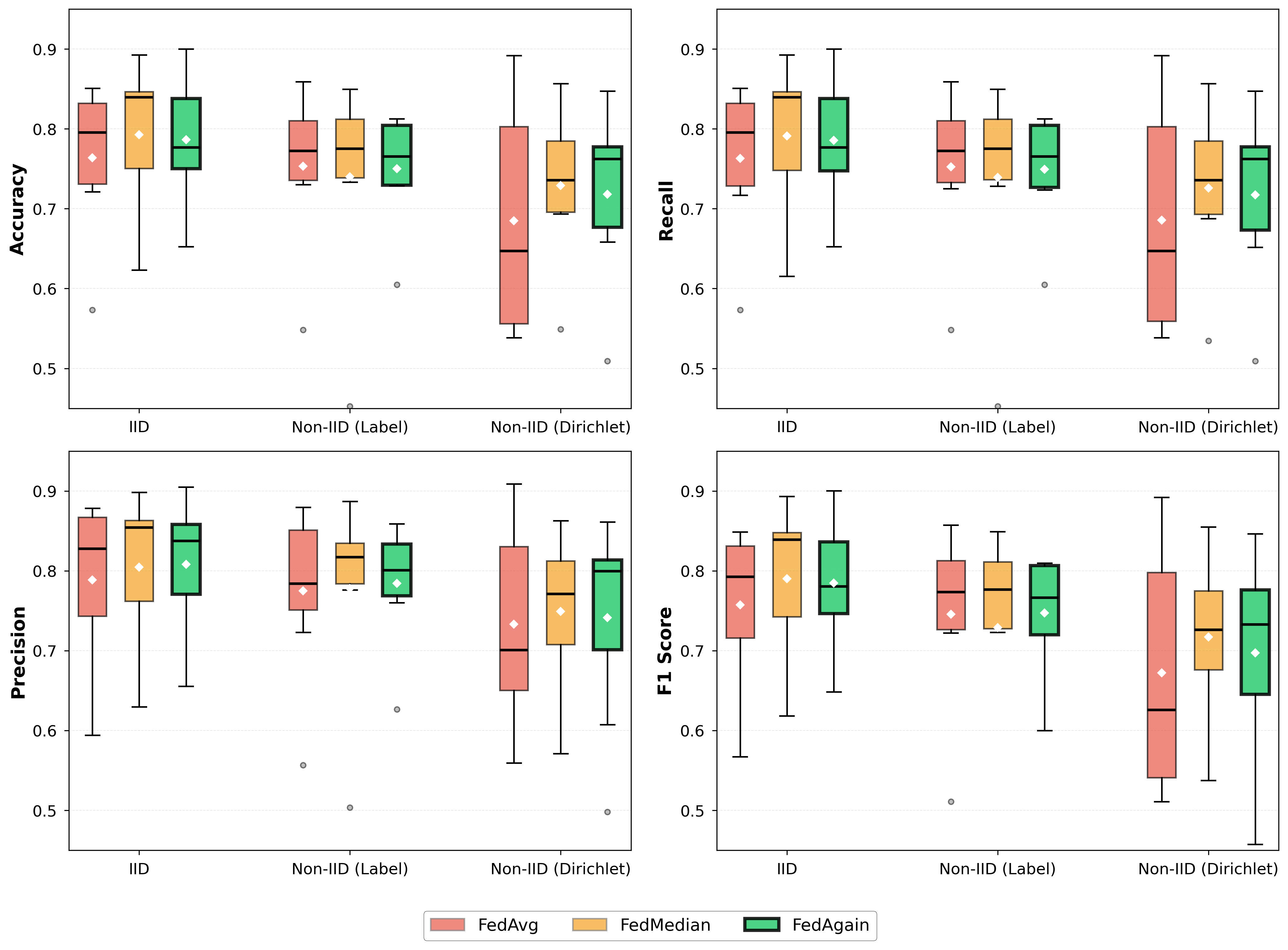}
\caption{
FedAgain consistently outperforms FedAvg across IID, Non-IID (Label), and Non-IID (Dirichlet) settings under 30\% client corruption (severity 3), with up to +4.8\% accuracy gain in the most heterogeneous scenario. It also shows lower variance and competitive performance against FedMedian, while adding trust-based weighting to better handle unreliable clients in realistic federated environments.
}
\label{fig:fedagain_vs_fedavg_distributions}
\end{figure*}
FedAvg, FedMedian, and FedAgain were evaluated under three data distribution paradigms to assess their robustness in realistic heterogeneous federated settings. All experiments were conducted on the seven kidney stone image sub-sets using 10 clients, 10 communication rounds, and 30\% corrupted clients at severity level 3.

Three heterogeneity regimes were considered: \emph{(i)} IID, where data were uniformly distributed across clients with balanced class representation; \emph{(ii)} non-IID Label Skew, where each client predominantly contained samples from a limited subset of classes, simulating specialized hospitals; and \emph{(iii)} non-IID Dirichlet, where data were partitioned using a Dirichlet distribution with concentration parameter $\alpha = 0.5$, producing both class imbalance and unequal client dataset sizes.

Figure \ref{fig:fedagain_vs_fedavg_distributions} presents box plots for accuracy, precision, recall, and F1-score across the three aggregation methods. Each box summarizes the median, mean, interquartile range, and outliers.

Under IID conditions, all methods achieved strong performance. FedMedian obtained the highest accuracy at $79.29 \pm 9.4\%$, followed by FedAgain at $78.65 \pm 8.1\%$ and FedAvg at $76.36 \pm 9.8\%$. FedAgain improved accuracy by $2.99\%$ over FedAvg while remaining within $0.80\%$ of FedMedian. It also achieved the highest precision ($80.79\%$) and the lowest variance, indicating that trust-based aggregation preserves performance while improving consistency under ideal conditions.

In the non-IID Label Skew setting, FedAgain achieved $75.00 \pm 7.3\%$ accuracy, comparable to FedAvg ($75.31 \pm 10.2\%$) and higher than FedMedian ($73.98 \pm 13.3\%$). Although FedAvg was marginally higher in accuracy, FedAgain surpassed FedMedian in both accuracy and F1-score ($74.71\%$ vs.\ $72.88\%$), while showing substantially lower variance. This reduced dispersion suggests that FedAgain provides more stable performance when client label distributions are highly skewed.

In the non-IID Dirichlet setting, which represents the most challenging heterogeneity scenario, FedAvg dropped to $68.50 \pm 14.8\%$ accuracy, showing the widest dispersion across metrics. FedMedian achieved the highest accuracy at $72.90 \pm 10.0\%$, followed closely by FedAgain at $71.81 \pm 11.1\%$. Both robust methods clearly outperformed FedAvg, with FedAgain improving accuracy by $4.82\%$. Although slightly below FedMedian, FedAgain remained competitive while preserving lower variance than FedAvg across all distributions, highlighting its robustness to the joint effects of client corruption and data heterogeneity.

\begin{figure*}[h]
\centering
\includegraphics[width=\linewidth]{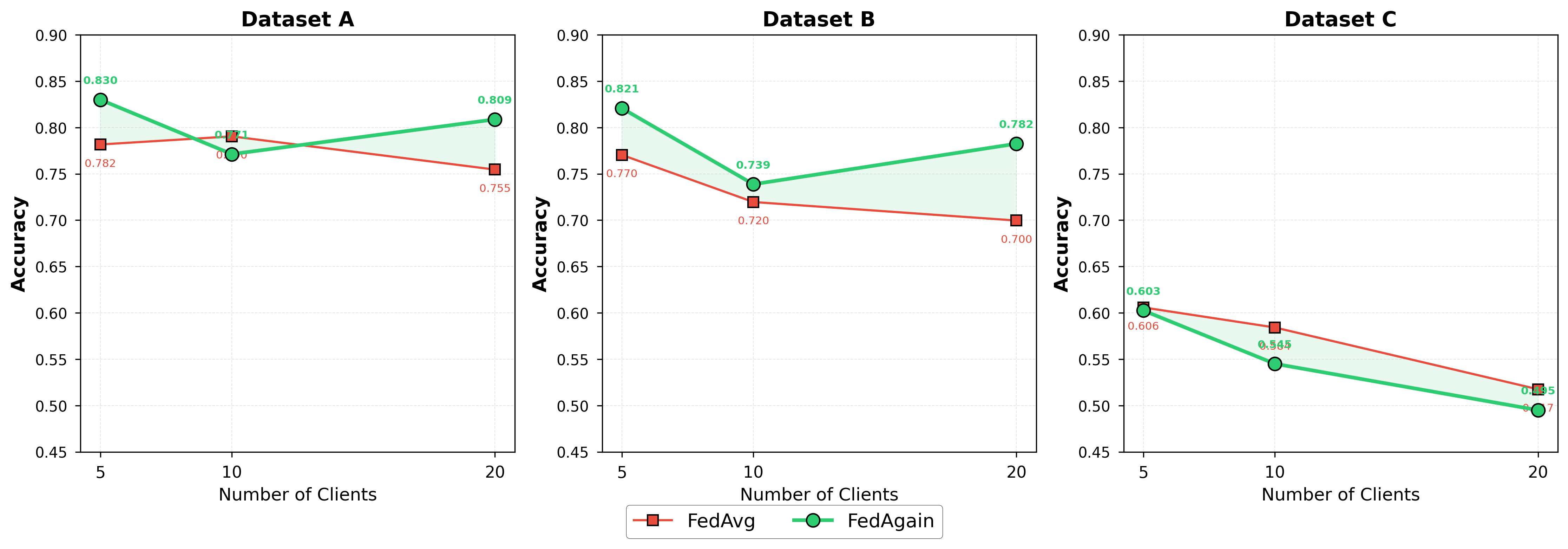}
\caption{
Scalability comparison of FedAvg and FedAgain as the number of participating clients increases from 5 to 20 under non-IID (Label Skew) data distribution with 30\% corrupted clients at severity level 3, evaluated on three kidney stone datasets. The green shaded region denotes the accuracy gap between both strategies. FedAgain (solid green line) consistently outperforms FedAvg (red line) on Datasets A and B, with the gap becoming more pronounced at 20 clients. Dataset C represents the most challenging classification setting, where both methods exhibit performance degradation as the number of clients increases.
}
\label{fig:fedagain_vs_fedavg_client_scaling}
\end{figure*}

\subsection{Performance Scalability Across Client Population Sizes}

Scalability is a key requirement for federated learning in healthcare, where adding more hospitals or clinical sites increases both data fragmentation and the likelihood of incorporating corrupted or low-quality participants. To examine this effect, we varied the number of participating clients from 5 to 10 and 20 while keeping the corruption ratio fixed at 30\%, corruption severity at level 3, and the data distribution under the non-IID Label Skew setting. This analysis focuses on FedAvg and FedAgain, since FedAvg remains the standard baseline in federated learning and provides the most relevant reference for assessing the scalability of a new aggregation strategy.

Figure \ref{fig:fedagain_vs_fedavg_client_scaling} illustrates the accuracy trends for three representative kidney stone datasets as the federation grows. On Dataset A, FedAgain achieved $83.00\%$ accuracy with 5 clients and retained $80.88\%$ with 20 clients, whereas FedAvg declined from $78.17\%$ to $75.46\%$. As a result, the performance margin in favor of FedAgain increased from $+4.83\%$ to $+5.42\%$. A more pronounced advantage is observed on Dataset B: FedAvg dropped from $77.04\%$ to $69.96\%$ as the number of clients increased, while FedAgain maintained stronger performance, decreasing only from $82.08\%$ to $78.25\%$. This yielded an improvement of $+5.04\%$ at 5 clients and $+8.29\%$ at 20 clients, indicating that the benefit of trust-based weighting becomes more evident as the federation grows and client heterogeneity intensifies.

Dataset C represents the most challenging classification setting, where both strategies deteriorated as the number of clients increased. FedAvg decreased from $60.58\%$ to $51.75\%$, while FedAgain dropped from $60.25\%$ to $49.50\%$. This behavior suggests that, for particularly difficult datasets, the combination of limited samples per class and stronger fragmentation across a larger number of clients can severely restrict the quality of local updates, regardless of the aggregation rule. However, the overall trend across the three datasets shows that FedAgain scales more gracefully than FedAvg in most cases. At 20 clients, FedAgain achieved an average accuracy of $69.54\%$, compared with $65.72\%$ for FedAvg, corresponding to a relative improvement of $+3.82$ percentage points. These findings suggest that trust-based aggregation offers a practical and scalable mechanism for preserving model quality as federated healthcare networks expand to include a larger and more diverse set of institutions.

\begin{figure*}[h]
\centering
\includegraphics[width=\linewidth]{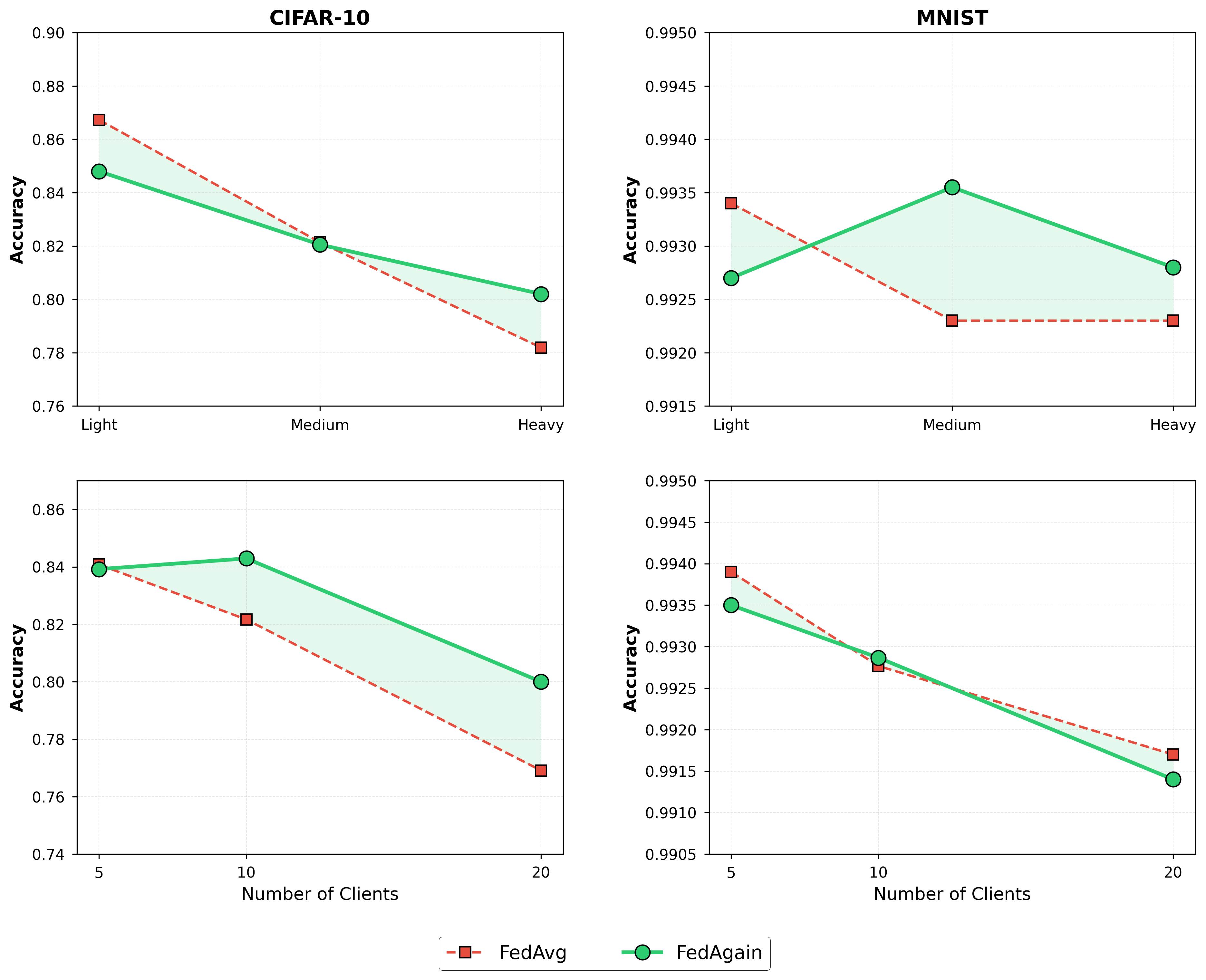}
\caption{
Ablation study on CIFAR-10 (left) and MNIST (right) comparing FedAgain with FedAvg under non-IID Label Skew conditions. \textbf{Top row:} Accuracy under progressively stronger corruption, with increasing fractions of corrupted clients and higher corruption severity. \textbf{Bottom row:} Accuracy as the number of participating clients increases from 5 to 20 under fixed corruption (30\%, severity 3). The shaded region denotes the performance gap between both methods. FedAgain shows a progressively larger advantage on CIFAR-10 as corruption intensifies and the federation scales, whereas both methods remain comparatively similar on MNIST, reflecting its lower task complexity.
}
\label{fig:cifar_mnist_simplified_analysis}
\end{figure*}

\subsection{Performance Validation on Standard Benchmarks}

To complement the medical imaging experiments, we conducted ablation studies on two standard benchmarks, CIFAR-10 and MNIST. These experiments isolate the effect of FedAgain's corruption-aware aggregation by directly comparing it with FedAvg, the canonical federated baseline. This design allows performance differences to be attributed specifically to the proposed aggregation mechanism, without the additional confounding factors introduced by alternative robust aggregation rules.

Figure \ref{fig:cifar_mnist_simplified_analysis} summarizes two complementary ablation settings. The top row evaluates robustness under progressively stronger corruption, where both the fraction of corrupted clients and the corruption severity increase across three scenarios: \textit{Light} (10\% corrupted clients, severity 3), \textit{Medium} (30\%, severity 4), and \textit{Heavy} (40\%, severity 5). The bottom row examines scalability by increasing the number of participating clients from 5 to 10 and 20 under fixed corruption conditions (30\% corrupted clients, severity 3). All experiments were conducted under a non-IID Label Skew partition to reflect realistic federated heterogeneity.

On CIFAR-10, FedAgain shows a progressively larger advantage as the setting becomes more challenging. Under light corruption, both methods perform similarly; however, as corruption intensifies, FedAvg degrades more sharply while FedAgain maintains higher accuracy, leading to an increasingly wider performance gap. The same trend is observed in the scalability analysis: although both methods are comparable at 5 clients, FedAgain preserves accuracy more effectively as the number of clients increases, indicating better robustness under growing heterogeneity and adversarial pressure.

On MNIST, both methods achieve near-ceiling accuracy ($>99.1\%$) across all evaluated conditions, which is consistent with the lower intrinsic difficulty of the dataset. The small difference between FedAvg and FedAgain indicates that the proposed method does not introduce degradation on simpler tasks, while its benefits become more evident on more challenging benchmarks such as CIFAR-10.

\section{Conclusion}

The results given in this paper show that FedAgain demonstrates robustness, scalability, and reliability across a wide range of federated learning configurations. Its adaptive trust-based aggregation mechanism favors rather constant performance under varying levels of corruption, heterogeneous data distributions, and diverse client populations. 

% By achieving 45\% \textcolor{blue}{<-from where comes this number?}lower performance variance compared to baseline methods, FedAgain offers predictable training behavior without requiring manual threshold tuning or complex Byzantine detection, making it highly practical for real-world clinical and multi-institutional deployments.

Across federation sizes, FedAgain consistently outperforms traditional averaging strategies such as FedAvg, with advantages that become more pronounced as the number of participating clients increases. This scalability reflects its ability to effectively isolate unreliable updates and leverage trustworthy client contributions, enabling convergence even when up to 40\% of participants are corrupted. The efficiency and adaptability of the algorithm ensure reliable performance in both small-scale controlled networks and large heterogeneous federations.

Overall, FedAgain provides a compelling foundation for robust federated learning in healthcare and beyond. Its performance-based weighting framework delivers consistent accuracy improvements, reduced variability, and greater resilience to adversarial and stochastic challenges. These properties establish FedAgain as a scalable and domain-agnostic solution capable of supporting reliable AI-driven collaboration between hospitals, research institutions, and broader distributed learning ecosystems.

%\subsection{Future Work}

Future research will extend the FedAgain robustness framework to additional domains and deployment conditions. Key directions include: i) integrating privacy-preserving mechanisms such as differential privacy and secure aggregation to improve compliance with medical data regulations, ii) exploring adaptive learning rate and communication-efficient extensions to improve scalability for large, multi-institutional federations, and iii) evaluating FedAgain’s performance in real-world clinical pipelines through longitudinal studies involving diverse imaging modalities and institutions. Further investigation of hybrid trust-weighted aggregation with personalized model adaptation may also strengthen performance in highly heterogeneous and evolving data environments.

\section*{Acknowledgments}
The authors wish to acknowledge the Mexican Secretaría de Ciencia, Humanidades, Tecnología e Innovación (Secihti) and CINVESTAV for their support in terms of postgraduate scholarships in this project, and the Data Science Hub at Tecnologico de Monterrey for their support on this project.
This work has been supported by Azure Sponsorship credits granted by Microsoft's AI for Good Research Lab through the AI for Health program.
The project was also supported by the French-Mexican Ecos-Nord grant 322537/M22M01.
We also gratefully acknowledge the support from the Google Explore Computer Science Research (CSR) Program for partially funding this project through the LATAM Undergraduate Research Program.
\bibliographystyle{cas-model2-names}
\bibliography{biblio}

@article{li2024role,
  title={Role of artificial intelligence in medical image analysis: A review of current trends and future directions},
  author={Li, Xin and Zhang, Lei and Yang, Jingsi and Teng, Fei},
  journal={Journal of Medical and Biological Engineering},
  volume={44},
  number={2},
  pages={231--243},
  year={2024},
  publisher={Springer}
}

@article{rajpurkar2017chexnet,
  title        = {CheXNet: Radiologist-Level Pneumonia Detection on Chest X-Rays with Deep Learning},
  author       = {Rajpurkar, Pranav and Irvin, Jeremy and Zhu, Kaylie and Yang, Brandon and Mehta, Hector and Duan, Dillon and Ding, Daisy and Bagul, Arjun and Langlotz, Curtis and Shpanskaya, Katie and others},
  journal      = {arXiv preprint},
  year         = {2017},
  url          = {https://arxiv.org/abs/1711.05225}
}

@article{ghorbani2020deeplearningechocardiogram,
  title        = {Deep learning interpretation of echocardiograms},
  author       = {Ghorbani, Amir and others},
  journal      = {npj Digital Medicine},
  volume       = {3},
  pages        = {10},
  year         = {2020},
  doi          = {10.1038/s41746-019-0216-8},
}

@article{esteva2017dermatologist,
  title        = {Dermatologist-level classification of skin cancer with deep neural networks},
  author       = {Esteva, Andre and Kuprel, Brett and Novoa, Roberto A. and Ko, Justin and Swetter, Susan M. and Blau, Helen M. and Thrun, Sebastian},
  journal      = {Nature},
  volume       = {542},
  number       = {7639},
  pages        = {115–118},
  year         = {2017},
  doi          = {10.1038/nature21056},
}

@INPROCEEDINGS{Amaouche2025,
  author={Amaouche, Meryem and Karrakchou, Ouassim and Ghogho, Mounir and Daul, Christian and Elghazzaly, Anouar and Alami, Mohamed and Ameur, Ahmed},
  booktitle={2025 IEEE 35th International Workshop on Machine Learning for Signal Processing (MLSP)}, 
  title={Cycle-Consistent Diffusion Model With Vessel-Aware Attention for Endoscopic Image Translation}, 
  year={2025},
  volume={},
  number={},
  pages={01-06},
  doi={10.1109/MLSP62443.2025.11204283}
}

@article{litjens2017survey,
  title        = {A Survey on Deep Learning in Medical Image Analysis},
  author       = {Litjens, Geert and Kooi, Thijs and Bejnordi, Babak E. and Setio, Arnaud A.A. and Ciompi, Francesco and Ghafoorian, Mohsen and van der Laak, Jeroen A.W.M. and van Ginneken, Bram and Sánchez, Clara I.},
  journal      = {Medical Image Analysis},
  volume       = {42},
  pages        = {60--88},
  year         = {2017},
  doi          = {10.1016/j.media.2017.07.005}
}

@article{liu2021advances,
  title        = {Advances in Deep Learning‐Based Medical Image Analysis},
  author       = {Liu, Xia and others},
  journal      = {Health Data Science},
  volume       = {2021},
  year         = {2021},
  doi         = {10.34133/2021/8786793}
}

@article{bian2025ai,
  title        = {Artificial intelligence in medical imaging: From task-specific to foundation models},
  author       = {Bian, Y.},
  journal      = {Chinese Medical Journal},
  volume       = {138},
  number       = {5},
  year         = {2025},
  doi          = {10.1097/CM9.0000000000003489}
}

@article{matus2020trustworthy,
  title        = {The role of explainability in creating trustworthy artificial intelligence for health care: a comprehensive survey of the terminology, design choices, and evaluation strategies},
  author       = {Markus, Aniek F. and Kors, Jan A. and Rijnbeek, Peter R.},
  journal      = {Artificial Intelligence in Medicine},
  volume       = {107},
  pages        = {101912},
  year         = {2020},
  doi          = {10.1016/j.artmed.2020.101912}
}

@article{purwono2025xai,
  title        = {Explainable Artificial Intelligence (XAI) in Medical Imaging: Techniques, Applications, Challenges, and Future Directions},
  author       = {Purwono, Arif and Wulandari, Annastasya N. E. and Nisa, Khoirun},
  journal      = {AMMS Journal – Medical Imaging},
  volume       = {1},
  number       = {1},
  pages        = {52–66},
  year         = {2025},
  url          = {https://doi.org/10.53623/amms.v1i1.692}
}

@article{xu2024interpretable,
  title        = {Explainability, transparency and black box challenges of AI in cardiovascular imaging},
  author       = {Xu, Hanhui and Shuttleworth, Kyle Michael James},
  journal      = {Egyptian Journal of Radiology and Nuclear Medicine},
  year         = {2024},
  pages        = {13:56},
  doi          = {10.1186/s43055-024-01356-2}
}

@misc{hendrycks2019benchmarking,
  title={Benchmarking neural network robustness to common corruptions and perturbations},
  author={Hendrycks, Dan and Dietterich, Thomas},
  note = {Preprint at \url{https://arxiv.org/abs/1903.12261}},
  year={2019}
}

@article{yoon2024domain,
  title        = {Domain Generalization for Medical Image Analysis: A Survey},
  author       = {Yoon, J. S. and Oh, K. and Shin, Y. and Mazurowski, M. A. and Suk, H.-I.},
  journal      = {arXiv preprint arXiv:2310.08598v2},
  year         = {2024},
  url          = {https://arxiv.org/abs/2310.08598}
}

@article{matta2024systematic,
  title={A systematic review of generalization research in medical image classification},
  author={Matta, Sarah and Lamard, Mathieu and Zhang, Philippe and Le Guilcher, Alexandre and Borderie, Laurent and Cochener, B{\'e}atrice and Quellec, Gwenol{\'e}},
  journal={Computers in biology and medicine},
  volume={183},
  pages={109256},
  year={2024},
  publisher={Elsevier}
}

@article{koccak2025bias,
  title={Bias in artificial intelligence for medical imaging: fundamentals, detection, avoidance, mitigation, challenges, ethics, and prospects},
  author={Ko{\c{c}}ak, Burak and Ponsiglione, Andrea and Stanzione, Arnaldo and Bluethgen, Christian and Santinha, Jo{\~a}o and Ugga, Lorenzo and Huisman, Merel and Klontzas, Michail E and Cannella, Roberto and Cuocolo, Renato},
  journal={Diagnostic and interventional radiology},
  volume={31},
  number={2},
  pages={75},
  year={2025}
}

@article{Zeteno2022Gastro,
title = {Optical biopsy mapping on endoscopic image mosaics with a marker-free probe},
journal = {Computers in Biology and Medicine},
volume = {143},
pages = {105234},
year = {2022},
doi = {https://doi.org/10.1016/j.compbiomed.2022.105234},
author = {Omar Zenteno and Dinh-Hoan Trinh and Sylvie Treuillet and Yves Lucas and Thomas Bazin and Dominique Lamarque and Christian Daul},
}

@INPROCEEDINGS{Espinosa2024Colon,
  author={Espinosa, Ricardo and Cerriteño, Javier and Gonzalez-Dominguez, Saul and Ochoa-Ruiz, Gilberto and Daul, Christian},
  booktitle={IEEE 37th International Symposium on Computer-Based Medical Systems (CBMS)}, 
  title={A deep learning-based image pre-processing pipeline for enhanced {3D} colon surface reconstruction robust to endoscopic illumination artifacts}, 
  year={2024},
  volume={},
  number={},
  pages={81-88},
  doi={10.1109/CBMS61543.2024.00022}
}

@article{Ali2016Bladder,
title = {Anisotropic motion estimation on edge preserving Riesz wavelets for robust video mosaicing},
journal = {Pattern Recognition},
volume = {51},
pages = {425-442},
year = {2016},

doi = {10.1016/j.patcog.2015.09.021},
author = {Sharib Ali and Christian Daul and Ernest Galbrun and François Guillemin and Walter Blondel}
}

@INPROCEEDINGS{Villalvazo2023KidneyStones,
  author={Villalvazo-Avila, Elias and Lopez-Tiro, Francisco and Rizo, Mariana and El-Beze, Jonathan and Hubert, Jacques and Gonzalez-Mendoza, Miguel and Ochoa-Ruiz, Gilberto and Daul, Christian},
  booktitle={IEEE 20th International Symposium on Biomedical Imaging (ISBI)}, 
  title={Improved Kidney Stone Recognition Through Attention and Multi-View Feature Fusion Strategies}, 
  year={2023},
  volume={},
  number={},
  pages={1-5},
  doi={10.1109/ISBI53787.2023.10230794}
}

@article{ali2022aiendoscopy,
  title        = {Where do we stand in AI for endoscopic image analysis? A review and meta-analysis},
  author       = {Ali, Sharib},
  journal      = {NPJ Digital Medicine},
  volume       = {5},
  number       = {1},
  pages        = {50},
  year         = {2022},
  doi          = {10.1038/s41746-022-00733-3}
}

@article{dretler1988laser,
  title={Laser lithotripsy: a review of 20 years of research and clinical applications},
  author={Dretler, Stephen P},
  journal={Lasers in surgery and medicine},
  volume={8},
  number={4},
  pages={341--356},
  year={1988},
  publisher={Wiley Online Library}
}

@article{elbeze2022evaluation,
  title        = {Evaluation and understanding of automated urinary stone recognition methods},
  author       = {El Beze, Jonathan and  Mazeaud, Charles and  Daul, Christian and  Ochoa‐Ruiz, Gilberto and  Daudon, Michel and Eschwège, Pascal and  Hubert, Jacques},
  journal      = {BJU international},
  volume       = {130},
  number       = {6},
  pages        = {786--798},
  year         = {2022},
  doi          = {10.1111/bju.15767}
}

@article{rieke2020future,
  title={The future of digital health with federated learning},
  author={Rieke, Nicola and Hancox, Jonny and Li, Wenqi and Milletar{\`i}, Fausto and Roth, Holger R and Albarqouni, Shadi and Bakas, Spyridon and Galtier, Mathieu N and Landman, Bennett A and Maier-Hein, Klaus and others},
  journal={NPJ Digital Medicine},
  volume={3},
  number={1},
  pages={119},
  year={2020}
}

@article{sheller2020federated,
  title={Federated learning in medicine: facilitating multi-institutional collaborations without sharing patient data},
  author={Sheller, Micah J and Edwards, Brandon and Reina, G Anthony and Martin, Jason and Pati, Sarthak and Kotrotsou, Aikaterini and Milchenko, Mikhail and Xu, Weilin and Marcus, Daniel and Colen, Rivka R and others},
  journal={Scientific reports},
  volume={10},
  pages={12598},
  year={2020},
  doi = {10.1038/s41598-020-69250-1}
}

@article{xu2021federated,
  title={Federated learning for healthcare informatics},
  author={Xu, Jie and Glicksberg, Benjamin S and Su, Chao and Walker, Peter and Bian, Jiang and Wang, Fei},
  journal={Journal of Healthcare Informatics Research},
  volume={5},
  number={1},
  pages={1--19},
  year={2021},
  publisher={Springer}
}

@article{antunes2022federated,
  title={Federated learning for healthcare: Systematic review and architecture proposal},
  author={Antunes, Rodolfo Stoffel and Andr{\'e} da Costa, Cristiano and K{\"u}derle, Arne and Yari, Imrana Abdullahi and Eskofier, Bj{\"o}rn},
  journal={ACM Transactions on Intelligent Systems and Technology (TIST)},
  volume={13},
  number={4},
  pages={1--23},
  year={2022},
  publisher={ACM New York, NY}
}

@article{li2020multi,
  title={Multi-institutional deep learning modeling without sharing patient data: A feasibility study on brain tumor segmentation},
  author={Li, Wenqi and Milletar{\`i}, Fausto and Xu, Dan and Rieke, Nicola and Hancox, Jonny and Zhu, Wenjia and Baust, Maximilian and Cheng, Yuyin and Ourselin, Sebastien and Cardoso, M Jorge and others},
  journal={Brainlesion: Glioma, Multiple Sclerosis, Stroke and Traumatic Brain Injuries},
  pages={132--144},
  year={2020},
  publisher={Springer}
}

@article{dou2021federated,
  title={Federated deep learning for detecting COVID-19 lung abnormalities in CT: a privacy-preserving multinational validation study},
  author={Dou, Qi and So, Theodore Y and Jiang, Min and Liu, Qi and Vardhanabhuti, Varut and Kaissis, Georgios and Li, Wenqi and Si, Weixin and Lee, Heung-Il and Yu, Ka Chun and others},
  journal={NPJ Digital Medicine},
  volume={4},
  number={1},
  pages={60},
  year={2021},
  publisher={Nature Publishing Group}
}

@article{guan2024federated,
  title={Federated learning for medical image analysis: A survey},
  author={Guan, Hao and Yap, Pew-Thian and Bozoki, Andrea and Liu, Mingxia},
  journal={Pattern Recognition},
  pages={110424},
  year={2024},
  publisher={Elsevier}
}

@article{novoa2023fast,
  title={Fast deep autoencoder for federated learning},
  author={Novoa-Paradela, David and Fontenla-Romero, Oscar and Guijarro-Berdi{\~n}as, Bertha},
  journal={Pattern Recognition},
  volume={143},
  pages={109805},
  year={2023},
  publisher={Elsevier}
}

@article{laridi2024enhanced,
  title={Enhanced federated anomaly detection through autoencoders using summary statistics-based thresholding},
  author={Laridi, Sofiane and Palmer, Gregory and Tam, Kam-Ming Mark},
  journal={Scientific Reports},
  volume={14},
  number={1},
  pages={26704},
  year={2024},
  publisher={Nature Publishing Group UK London}
}

@article{kaissis2020secure,
  title={Secure, privacy-preserving and federated machine learning in medical imaging},
  author={Kaissis, Georgios A and Makowski, Marcus R and R{\"u}ckert, Daniel and Braren, Rickmer F},
  journal={Nature Machine Intelligence},
  volume={2},
  number={6},
  pages={305--311},
  year={2020},
  publisher={Nature Publishing Group UK London}
}

@inproceedings{reyes2025robust,
  title={Robust Federated Anomaly Detection Using Dual-Signal Autoencoders: Application to Kidney Stone Identification in Ureteroscopy},
  author={Reyes-Amezcua, Ivan and Lopez-Tiro, Francisco and Larose, Cl{\'e}ment and Daul, Christian and Mendez-Vazquez, Andres and Ochoa-Ruiz, Gilberto},
  booktitle={MICCAI Workshop on Data Engineering in Medical Imaging},
  pages={125--135},
  year={2025},
  organization={Springer}
}

@inproceedings{lopez2021assessing,
  title={Assessing deep learning methods for the identification of kidney stones in endoscopic images},
  author={Lopez, Francisco and Varelo, Andres and Hinojosa, Oscar and Mendez, Mauricio and Trinh, Dinh-Hoan and ElBeze, Yonathan and Hubert, Jacques and Estrade, Vincent and Gonzalez, Miguel and Ochoa, Gilberto and Daul, Christian},
  booktitle={43rd Annual International Conference of the IEEE Engineering in Medicine \& Biology Society (EMBC)},
  pages={2778--2781},
  year={2021},
  organization={IEEE}
}

@article{lopez2024vivo,
  title={On the in vivo recognition of kidney stones using machine learning},
  author={Lopez-Tiro, Francisco and Estrade, Vincent and Hubert, Jacques and Flores-Araiza, Daniel and Gonzalez-Mendoza, Miguel and Ochoa, Gilberto and Daul, Christian},
  journal={IEEE Access},
  volume={12},
  pages={10736--10759},
  year={2024},
  publisher={IEEE}
}

@INPROCEEDINGS{Trinh2017IlluminationChanges,
  author={Trinh, Dinh-Hoan and Blondel, Walter and Daul, Christian},
  booktitle={IEEE International Conference on Image Processing (ICIP)}, 
  title={A general form of illumination-invariant descriptors in variational optical flow estimation}, 
  year={2017},
  volume={},
  number={},
  pages={2533-2537},
  doi={10.1109/ICIP.2017.8296739}
}

@INPROCEEDINGS{GarciaVega2023ExposureCorrection,
  author={García-Vega, Axel and Espinosa, Ricardo and Ramírez-Guzmán, Luis and Bazin, Thomas and Falcón-Morales, Luis and Ochoa-Ruiz, Gilberto and Lamarque, Dominique and Daul, Christian},
  booktitle={IEEE 20th International Symposium on Biomedical Imaging (ISBI)}, 
  title={Multi-Scale Structural-aware Exposure Correction for Endoscopic Imaging}, 
  year={2023},
  volume={},
  number={},
  pages={1-5},
  doi={10.1109/ISBI53787.2023.10230724}
}

@INPROCEEDINGS{Weibel2012BladderCartography,
  author={Weibel, Thomas and Daul, Christian and Wolf, Didier and Rösch, Ronald},
  booktitle={Proceedings of the 21st International Conference on Pattern Recognition (ICPR)}, 
  title={Contrast-enhancing seam detection and blending using graph cuts}, 
  year={2012},
  volume={},
  number={},
  pages={2732-2735}
}

@misc{wang2023survey,
  title = "A Survey on the Robustness of Computer Vision Models against Common Corruptions",
  author = "Shunxin Wang and Raymond Veldhuis and Christoph Brune and Nicola Strisciuglio",
  note = {Preprint at \url{https://arxiv.org/abs/2305.06024}},
  year = "2023"
}

@misc{fang2023robust,
  title={Robust heterogeneous federated learning under data corruption},
  author={Fang, Xiuwen and Ye, Mang and Yang, Xiyuan},
  note={Paper presented at the IEEE/CVF International Conference on Computer Vision},
  year={2023}
}

@article{pillutla2022robust,
  title={Robust aggregation for federated learning},
  author={Pillutla, Krishna and Kakade, Sham M and Harchaoui, Zaid},
  journal={IEEE Transactions on Signal Processing},
  volume={70},
  pages={1142--1154},
  year={2022},
  publisher={IEEE}
}

@inproceedings{mcmahan2017communication,
  title={Communication-efficient learning of deep networks from decentralized data},
  author={McMahan, Brendan and Moore, Eider and Ramage, Daniel and Hampson, Seth and y Arcas, Blaise Aguera},
  booktitle={Artificial intelligence and statistics},
  pages={1273--1282},
  year={2017},
  organization={PMLR}
}

@article{ng2021federated,
  title={Federated learning: a collaborative effort to achieve better medical imaging models for individual sites that have small labelled datasets},
  author={Ng, Dianwen and Lan, Xiang and Min-Szu Yao, Melissa and  Chan, Wing and  Feng, P. Mengling},
  journal={Quantitative Imaging in Medicine and Surgery},
  volume={11},
  pages={852},
  year={2021},
}

@article{li2020federated,
  title={Federated optimization in heterogeneous networks},
  author={Li, Tian and Sahu, Anit Kumar and Zaheer, Manzil and Sanjabi, Maziar and Talwalkar, Ameet and Smith, Virginia},
  journal={Proceedings of Machine learning and systems},
  volume={2},
  pages={429--450},
  year={2020}
}

@article{kairouz2021advances,
  title={Advances and open problems in federated learning},
  author={Kairouz, Peter and McMahan, H Brendan and Avent, Brendan and Bellet, Aur{\'e}lien and Bennis, Mehdi and Bhagoji, Arjun Nitin and Bonawitz, Kallista and Charles, Zachary and Cormode, Graham and Cummings, Rachel and others},
  journal={Foundations and trends{\textregistered} in machine learning},
  volume={14},
  number={1--2},
  pages={1--210},
  year={2021},
  publisher={Now Publishers, Inc.}
}

@article{li2019survey,
  title        = {A Survey on Federated Learning: Challenges, Methods, and Future Directions},
  author       = {Li, Tian and Sahu, Anit Kumar and Talwalkar, Ameet and Smith, Virginia},
  journal      = {IEEE Transactions on Knowledge and Data Engineering},
  volume       = {34},
  number       = {12},
  pages        = {727–746},
  year         = {2022},
  doi          = {10.1109/TKDE.2020.2982121}
}

@inproceedings{yin2018byzantine,
  author    = {Yin, Dong and Chen, Yudong and Kannan, R. and Bartlett, Peter L.},
  title     = {Byzantine‐Robust Distributed Learning: Towards Optimal Statistical Rates},
  booktitle = {Proceedings of the 35th International Conference on Machine Learning (ICML)},
  year      = {2018},
  pages     = {5650–5659}
}

@inproceedings{guerraoui2018hidden,
  title={The hidden vulnerability of distributed learning in byzantium},
  author={Guerraoui, Rachid and Rouault, S{\'e}bastien and others},
  booktitle={International conference on machine learning},
  pages={3521--3530},
  year={2018},
  organization={PMLR}
}

@article{hsu2019nonIID,
  author    = {Hsu, Tyler M.H. and Qi, H. and Brown, Michael},
  title     = {Measuring the Effects of Non‐Identical Data Distribution for Federated Visual Classification},
  journal   = {arXiv preprint arXiv:1909.06335},
  year      = {2019}
}

@article{yao2022lightfed,
  title={LightFed: Lightweight federated learning with class imbalance-aware aggregation},
  author={Yao, Li and Li, Qinbin and Lin, Zhiyong and Jiang, Xiaohua and Chan, Keith C C},
  journal={IEEE Transactions on Neural Networks and Learning Systems},
  year={2022},
  publisher={IEEE}
}

@article{kasidas2004renal,
  title={Renal stone analysis: why and how?},
  author={Kasidas, GP and Samuell, CT and Weir, TB},
  journal={Annals of clinical biochemistry},
  volume={41},
  number={2},
  pages={91--97},
  year={2004},
  publisher={SAGE Publications Sage UK: London, England}
}

@article{cloutier2015kidney,
  title={Kidney stone analysis:“Give me your stone, I will tell you who you are!”},
  author={Cloutier, Jonathan and Villa, Luca and Traxer, Olivier and Daudon, Michel},
  journal={World journal of urology},
  volume={33},
  number={2},
  pages={157--169},
  year={2015},
  publisher={Springer}
}

@article{daudon2016comprehensive,
  title={Comprehensive morpho-constitutional analysis of urinary stones improves etiological diagnosis and therapeutic strategy of nephrolithiasis},
  author={Daudon, Michel and Dessombz, Arnaud and Frochot, Vincent and Letavernier, Emmanuel and Haymann, Jean-Philippe and Jungers, Paul and Bazin, Dominique},
  journal={Comptes Rendus Chimie},
  volume={19},
  number={11-12},
  pages={1470--1491},
  year={2016},
  publisher={Elsevier}
}

@article{corrales2021classification,
  title={Classification of stones according to Michel Daudon: a narrative review},
  author={Corrales, Mariela and Doizi, Steeve and Barghouthy, Yazeed and Traxer, Olivier and Daudon, Michel},
  journal={European Urology Focus},
  volume={7},
  number={1},
  pages={13--21},
  year={2021}
}

@article{estrade2021ESR,
  title        = {Toward improved endoscopic examination of urinary stones: a concordance study between endoscopic digital pictures vs microscopy.},
  author       = {Estrade, Vincent and Denis de Senneville, Baudoin and Meria, Paul and  Almeras, Christophe and Bladou, Frank and Bernhard, Jean-Christophe and Robert, Grégoire and Traxer, Olivier and Daudon, Michel},
  journal      = {BJU international},
  volume       = {128},
  number       = {6},
  pages        = {319--330},
  year         = {2021},
  doi          = {10.1111/bju.15312 }
}

@article{Turcotte2025comprehensiveAnalysis,
title = {Comprehensive analysis of 55,213 stones: understanding common morphological associations advances endoscopic stone recognition and {AI} integration},
journal = {World Journal of Urology},
volume = {43},
number = {630},
pages = {},
year = {2025},
doi = {10.1007/s00345-025-05961-2},
author = {Turcotte, Bruno and Bernhard, Jean-Christophe and Bladou, Franck and Chicaud, Marie and Robert, Grégoire and Denis de Senneville, Baudouin  and  Daudon, Michel and Estrade, Vincent}
}

@article{Gonzalez2024MetricLearning,
title = {A metric learning approach for endoscopic kidney stone identification},
journal = {Expert Systems with Applications},
volume = {255},
pages = {124711},
year = {2024},
issn = {0957-4174},
doi = {10.1016/j.eswa.2024.124711},
author = {Jorge Gonzalez-Zapata and Francisco Lopez-Tiro and Elias Villalvazo-Avila and Daniel Flores-Araiza and Jacques Hubert and Gilberto Ochoa-Ruiz and Christian Daul and Andres Mendez-Vazquez}
}

@article{Florez2025ProtoParts,
title = {Improving prototypical parts abstraction for case-based reasoning explanations designed for the kidney stone type recognition},
journal = {Artificial Intelligence in Medicine},
volume = {170},
pages = {103266},
year = {2025},
doi = {10.1016/j.artmed.2025.103266},
author = {Daniel Flores-Araiza and Francisco Lopez-Tiro and Clément Larose and Salvador Hinojosa and Andres Mendez-Vazquez and Miguel Gonzalez-Mendoza and Gilberto Ochoa-Ruiz and Christian Daul}
}

@article{serrat2017mystone,
  title={myStone: A system for automatic kidney stone classification},
  author={Serrat, Joan and Lumbreras, Felipe and Blanco, Francisco and Valiente, Manuel and L{\'o}pez-Mesas, Montserrat},
  journal={Expert Systems with Applications},
  volume={89},
  pages={41--51},
  year={2017},
  publisher={Elsevier}
}

@article{deng2012mnist,
  title        = {The MNIST database of handwritten digit images for machine learning research},
  author       = {Deng, Li},
  journal      = {IEEE Signal Processing Magazine},
  volume       = {29},
  number       = {6},
  pages        = {141--142},
  year         = {2012},
  publisher    = {IEEE}
}

@techreport{Krizhevsky09learningmultiple,
  author       = {Krizhevsky, Alex and Nair, Vinod and Hinton, Geoffrey},
  title        = {Learning multiple layers of features from tiny images},
  institution  = {University of Toronto, Department of Computer Science},
  year         = {2009},
  note         = {Technical report},
  url          = {https://www.cs.toronto.edu/~kriz/cifar.html}
}

@inproceedings{liu2023byzantine,
  title={Byzantine-robust learning on heterogeneous data via gradient splitting},
  author={Liu, Yuchen and Chen, Chen and Lyu, Lingjuan and Wu, Fangzhao and Wu, Sai and Chen, Gang},
  booktitle={International Conference on Machine Learning},
  pages={21404--21425},
  year={2023},
  organization={PMLR}
}

@inproceedings{reguieg2023comparative,
  title={A comparative evaluation of fedavg and per-fedavg algorithms for dirichlet distributed heterogeneous data},
  author={Reguieg, Hamza and El Hanjri, Mohammed and El Kamili, Mohamed and Kobbane, Abdellatif},
  booktitle={2023 10th International Conference on Wireless Networks and Mobile Communications (WINCOM)},
  pages={1--6},
  year={2023},
  organization={IEEE}
}

@inproceedings{reyes2024leveraging,
  title={Leveraging Pre-trained Models for Robust Federated Learning for Kidney Stone Type Recognition},
  author={Reyes-Amezcua, Ivan and Rojas-Ruiz, Michael and Ochoa-Ruiz, Gilberto and Mendez-Vazquez, Andres and Daul, Christian},
  booktitle={Mexican International Conference on Artificial Intelligence},
  pages={168--181},
  year={2024},
  organization={Springer}
}

@misc{fang2022robust,
  title={Robust federated learning with noisy and heterogeneous clients},
  author={Fang, Xiuwen and Ye, Mang},
  booktitle={Paper presented at the IEEE/CVF Conference on Computer Vision and Pattern Recognition},
  year={2022}
}

\end{document}